\documentclass[journal]{IEEEtran}

\usepackage{amsfonts}
\usepackage{amssymb}
\usepackage{stfloats}
\usepackage{graphicx}

\usepackage{amsmath}
\usepackage{array}
\usepackage{epstopdf}
\usepackage{amsthm}

\usepackage[ruled,vlined]{algorithm2e}

\usepackage[noend]{algpseudocode}
\usepackage{amsmath}
\usepackage{graphics}
\usepackage{epsfig}
\usepackage{float}
\usepackage{booktabs}
\usepackage{multirow}

\usepackage{bm}
\usepackage{upgreek}
\usepackage{caption}
\usepackage{subcaption}
\usepackage{tabularx}
\usepackage{cite}
\usepackage{epstopdf}
\newlength\myindent
\usepackage[usenames]{color}

\usepackage{xcolor}

\newcommand\hlbreakable[1]{\textcolor{black}{#1}}
\usepackage[switch]{lineno}
\usepackage{mathtools}

\begin{document}

\title{Coop-WD: Cooperative Perception with Weighting and Denoising for Robust V2V Communication}

\author{
    Chenguang Liu, Jianjun Chen, Yunfei Chen, {\em{Fellow,~IEEE}}, Yubei He, Zhuangkun Wei\\Hongjian Sun, {\em{Senior~Member,~IEEE}}, Haiyan Lu, {\em{Senior~Member,~IEEE}}, and Qi Hao, {\em{Senior~Member,~IEEE}}
    \thanks{This work was supported in part by the Engineering and Physical Sciences Research Council: Communications Hub For Empowering Distributed ClouD Computing Applications And Research (CHEDDAR) (EP/X040518/1 and EP/Y037421/1), the Shenzhen Fundamental Research Program(JCYJ20220818103006012, KJZD20231023092600001), and National Natural Science Foundation of China(62261160654).}
    \thanks{Chenguang~Liu, Yunfei~Chen, Yubei~He, Zhuangkun~Wei, and Hongjian~Sun are with the Department of Engineering, Durham University, DH1 3LE, UK.(Email: chenguang.liu;yunfei.chen;yubei.he;zhuangkun.wei; hongjian.sun@durham.ac.uk)}
    \thanks{Jianjun Chen and Haiyan Lu are with the Faculty of Engineering and Information Technology, University of Technology, Sydney, Australia. (Email: jianjun.chen@student.uts.edu.au and haiyan.lu@uts.edu.au)} 
    \thanks{Qi Hao is with the Research Institute of Trustworthy Autonomous Systems and the Department of Computer Science and Engineering, Southern University of Science and Technology, Shenzhen, China. (Email: hao.q@sustech.edu.cn)} 
    \thanks{$^{*}$Corresponding author: Qi Hao. }% <-this % stops a space
    % \thanks{$^{1}$Durham University, Durham, UK.}%
    % \thanks{$^{2}$Southern University of Science and Technology, Shenzhen, China.}%
    % \thanks{$^{3}$University of Technology Sydney, Sydney, Australia.}%    
}

\maketitle

\begin{abstract}
Cooperative perception, leveraging shared information from multiple vehicles via vehicle-to-vehicle (V2V) communication, plays a vital role in autonomous driving to alleviate the limitation of single-vehicle perception. Existing works have explored the effects of V2V communication impairments on perception precision, but they lack generalization to different levels of impairments. In this work, we propose a joint weighting and denoising framework, \textit{Coop-WD}, to enhance cooperative perception subject to V2V channel impairments. In this framework, the self-supervised contrastive model and the conditional diffusion probabilistic model are adopted hierarchically for vehicle-level and pixel-level feature enhancement. An efficient variant, \textit{Coop-WD-eco}, is also proposed to selectively deactivate denoising to reduce processing overhead. Rician fading, non-stationarity, and time-varying distortion are considered. Simulation results demonstrate that the proposed \textit{Coop-WD} outperforms conventional benchmarks in all cases considered. Moreover, \textit{Coop-WD} is validated as a plug-in module with different fusion schemes, demonstrating its generalization capability for communication-aware cooperative perception enhancement. Qualitative analysis with visual examples further proves the superiority of our proposed method. The proposed \textit{Coop-WD-eco} achieves up to 50\% reduction in computational cost under severe distortion while maintaining comparable accuracy.

\end{abstract}


\section{Introduction}\label{sec1}
Perception of the dynamic environment is vital for proactive collision avoidance and route planning in autonomous driving. Emergent technologies, such as LiDAR and artificial intelligence (AI), have facilitated the development of 3D object detection in autonomous vehicles \cite{pointpillars}\cite{shi2020pv}. However, these sensors can be limited by their physical capabilities, including restricted detection range and scan frequency, which may prevent a single autonomous vehicle from accurately detecting occluded or distant objects. To address this, cooperative perception enabled by vehicular communications has been widely used by fusing sensed information from multiple connected autonomous vehicles (CAVs) to obtain wider viewpoints and enhance detection \cite{f-cooper,openv2v,v2vnet,v2x-vit, cobevt, when2com, where2comm,who2com}. This collaborative framework heavily relies on the reliability of vehicle-to-vehicle (V2V) communications, as the non-stationarity of the V2V channel due to moving scatterers can impact the information sharing among CAVs. Meanwhile, with the advances of the sixth-generation (6G) wireless communication\cite{10054381, 8869705}, AI has emerged as one of the key enablers for enhancing the robustness against noise and interference through end-to-end transceiver design in many applications, such as signal detection \cite{8054694, 9733260} and semantic communications \cite{9398576, 10345474}. Therefore, incorporating AI-enhanced communications in cooperative perception is promising for ensuring system robustness in cooperative perception systems, and this has attracted significant research interest. 

Based on the effective 3D detection backbones \cite{pointpillars,shi2020pv}, several studies have explored various collaborative fusion schemes for cooperative perception, including early fusion (sharing raw point clouds) \cite{early1,early2}, intermediate fusion (sharing intermediate feature representations) \cite{f-cooper,openv2v,v2vnet,v2x-vit, cobevt, when2com, where2comm,who2com}, and late fusion (sharing detection outcome) \cite{late1,late2}. These works aim to optimize the trade-offs between bandwidth utilization and detection accuracy, often assuming ideal communications. Recent works have shifted toward more realistic scenarios, considering various factors, such as communication delays \cite{v2x-vit}, information loss \cite{lossy}, and pose inaccuracies\cite{v2vnet,who2com}. Furthermore, the studies in \cite{guang1, guang2} demonstrated that learning-based communication systems can be effectively integrated into cooperative perception frameworks based on various detection backbones and fusion methods. This integration allows realistic channel impairments to be considered in the cooperative perception framework, including channel attenuation, path loss, and multi-path effects. However, a more sophisticated channel model that accounts for the non-stationarity of V2V channels, particularly the time-varying distortions caused by vehicle speed and moving scatterers, has yet to be investigated.

In a realistic cooperative perception system, shared information is designed to provide complementary features for better viewpoints; however, when inevitable channel impairments occur in V2V communication, the shared information can compromise the detection due to channel distortion. Although previous works have utilized both supervised end-to-end training \cite{guang1} and self-supervised learning \cite{guang2} to mitigate the adverse effects of channel impairments, several challenges remain. Specifically, supervised distortion-in-the-loop training \cite{guang1} allows fusion networks, such as attentive fusion and V2VNet, to learn from the distorted shared information, thereby improving detection performance under channel impairments. However, it lacks generalization to varying communication environments, as it struggles to function with varying noise levels, path loss factors, or other dynamic channel conditions. On the other hand, self-supervised learning offers a flexible approach to filtering out distorted information due to severe channel impairments. However, the existing methods operate only at the CAV level and cannot address pixel-level feature maps. Therefore, developing an approach that can flexibly adapt to varying V2V communication channels is essential to enhance the robustness of cooperative perception and address pixel-level feature maps.

Motivated by these existing challenges, this work will focus on improving the robustness of cooperative perception considering the dynamic and non-stationary nature of V2V communication. The contributions can be summarized as:
\begin{enumerate}
    \item To alleviate the performance degradation due to V2V communication impairments, we propose a joint weighting and denoising framework, \textit{Coop-WD}, to improve the robustness of cooperative perception by sharing intermediate features. A denoising algorithm based on a diffusion probabilistic model is developed to tackle pixel-level feature distortion. Additionally, we propose an efficient variant, \textit{Coop-WD-eco}, to selectively disable denoising module to reduce computational cost.
    \item Unlike the previous works that have neglected uncertainty and stochastic variability in V2V communication, a non-stationary V2V communication channel model is used to validate our method for a realistic V2V communication channel setting. Different channel conditions, noise levels, path loss factors, and time-varying distortions are considered to validate our approach. 
    \item Simulation results demonstrate that the proposed \textit{Coop-WD} framework consistently outperforms conventional benchmarks across all channel conditions and distortion levels, including simulated Rician fading, realistic WINNER II, and non-stationary V2V channels. An ablation study with quantitative results is conducted to evaluate the individual contributions of each module, alongside comparisons with baseline models. Additionally, qualitative analysis further validates that the proposed approach outperforms benchmarks that utilize only a weighting or denoising module.
\end{enumerate}

\section{Related work}

\subsection{Cooperative perception} 
Conventional single CAV perception faces inherent limitations, such as occlusion and limited field-of-view, which can result in inaccurate detection. To address these limitations, cooperative perception has garnered increasing interest in recent years.

The pioneer of early fusion was Cooper \cite{cooper}, which was proposed to conduct a study on point cloud level fusion for enhanced detection ability. The work in \cite{early1} proposed a central system to fuse point cloud data from multiple infrastructure sensors. Late fusion \cite{late1,late2,guang1} is another practical technique to achieve cooperative perception where the perception outputs of CAVs are shared to create the final results. The work in \cite{late1} proposed a two-layer architecture that handles object detection and tracking, where the information matrix fusion (IMF) algorithm was adopted to fuse the received data. The authors in \cite{late2} proposed a peer-to-peer architecture in which other CAVs transmitted the outputs of 3D bounding boxes to the ego vehicle. A new late fusion scheme was proposed in \cite{guang1} to leverage intermediate features in late fusion to enhance the robustness against wireless attenuation. 

For intermediate fusion, F-Cooper \cite{f-cooper} proposed two feature fusion schemes (\textit{i.e.}, voxel feature fusion and spatial feature fusion) to achieve efficient fusion under limited network bandwidth. V2VNet \cite{v2vnet} introduced a spatially-aware graph neural network (GNN) for joint perception and motion forecasting. This approach enhanced the detection accuracy by sharing intermediate features and utilizing spatial information from nearby CAVs. The work in \cite{openv2v} proposed an attention-based fusion algorithm to aggregate data from multiple vehicles. In addition, V2X-ViT \cite{v2x-vit} utilized a transformer-based architecture, combining heterogeneous multi-agent self-attention (HMSA) and multi-scale window attention (MSwin) to aggregate vehicle and infrastructure data, and address localization errors. \hlbreakable{Moreover, V2VFormer~\cite{10398509} introduced transformer-based spatial and channel-wise attention to recalibrate feature importance across CAVs, and V2VFormer++~\cite{10265751} extended this line of work with multi-modal camera-LiDAR fusion and a global-local transformer module for dynamic feature aggregation. More recently, CUDA-X~\cite{10891961} studied unsupervised domain adaptation for V2X collaborative perception through knowledge transfer and alignment, improving robustness in cross-domain deployment. These methods focus mainly on feature aggregation, multi-modal fusion, or domain adaptation, while the effect of non-ideal V2V wireless channels on transmitted intermediate features remains insufficiently studied.}

\subsection{Cooperative perception with V2V communication} 
Cooperative perception with V2V communication involves information exchange among CAVs via wireless channels. Thus, communication factors like latency, bandwidth, and channel distortion should be considered to maintain reliable and effective perception.

SyncNet, a latency-aware collaborative perception system, was introduced in~\cite{syncnet} to actively synchronize perceptual features shared among multiple CAVs. To reduce communication overhead in cooperative perception using V2V communication, FPV-RCNN~\cite{keypoints} introduced a keypoint-based deep feature fusion framework to reduce the shared data. Who2com \cite{who2com} introduced a learnable handshake communication mechanism, where CAVs selectively communicate based on relevance scores to reduce bandwidth utilization. Where2comm \cite{where2comm} proposed spatial confidence maps to allow CAVs to share only spatially sparse but perceptually critical information, thus avoiding the transmission of irrelevant information. Furthermore, ROBOSAC \cite{amongus} proposed an attacker-agnostic, sampling-based defense to enhance the cooperative perception with V2V communication under adversarial attacks and communication noise.

Most of these research works focus on reducing bandwidth usage and communications delay without considering impariments in wireless channels. Only a few works have explored cooperative perception under realistic communication channels. For instance, \cite{guang1} proposed a new late fusion scheme and a CNN-based autoencoder (AE) to account for Rician fading and path loss. \cite{guang2} developed a self-supervised adaptive weighting model to mitigate different levels of channel distortion. In \cite{lossy}, a lossy communication-aware repair network was combined with a V2V attention module to improve the resilience when there are information loss. However, V2V communication models have unique challenges, such as Doppler shifts \cite{doppler} due to moving CAVs, multi-path effects \cite{multipath} from static or dynamic objects, and other time-varying disturbances. These factors have not been investigated in the above works on cooperative perception with V2V communications.

\subsection{Diffusion models in wireless communication} 
Diffusion models have achieved unprecedented success in artificial intelligence-generated content (AIGC), such as image super-resolution \cite{10681246, diff_sr3}, image recovery \cite{diff_image}, and audio enhancement \cite{cdiffse}.

Due to the robust recovery and denoising capabilities of diffusion models, recent studies have begun exploring their applications in wireless communications. In \cite{diff_cddm}, a channel-denoising diffusion model was proposed to reduce noise and enhance received signals in wireless image transmission. In \cite{10816175}, conditional denoising diffusion probabilistic models were proposed for data reconstruction in wireless image transmission with hardware impairments. The work in \cite{10674003} proposed to use a diffusion model to remove the impact of hardware impairments, outdated CSI, and multi-user interference in cell-free massive MIMO systems. The authors in \cite{diff_bit} proposed a diffusion-guided framework for semantic communication that synthesizes high-quality images and preserves semantic content in noisy wireless environments. In \cite{diff_commin}, invertible neural networks were integrated with diffusion models to enhance joint source-channel coding for wireless image transmission, improving the semantic and perceptual fidelity of images in noisy wireless environments. These works have applied the diffusion models to wireless communications for image transmission. However, to the best of our knowledge, they have not been explored to address the information distortion due to channel impairments in practical V2V communications for cooperative perception.

\section{System Model}\label{sec2}
\begin{figure}[!h] 
  \centering
  \includegraphics[width=3.2in]{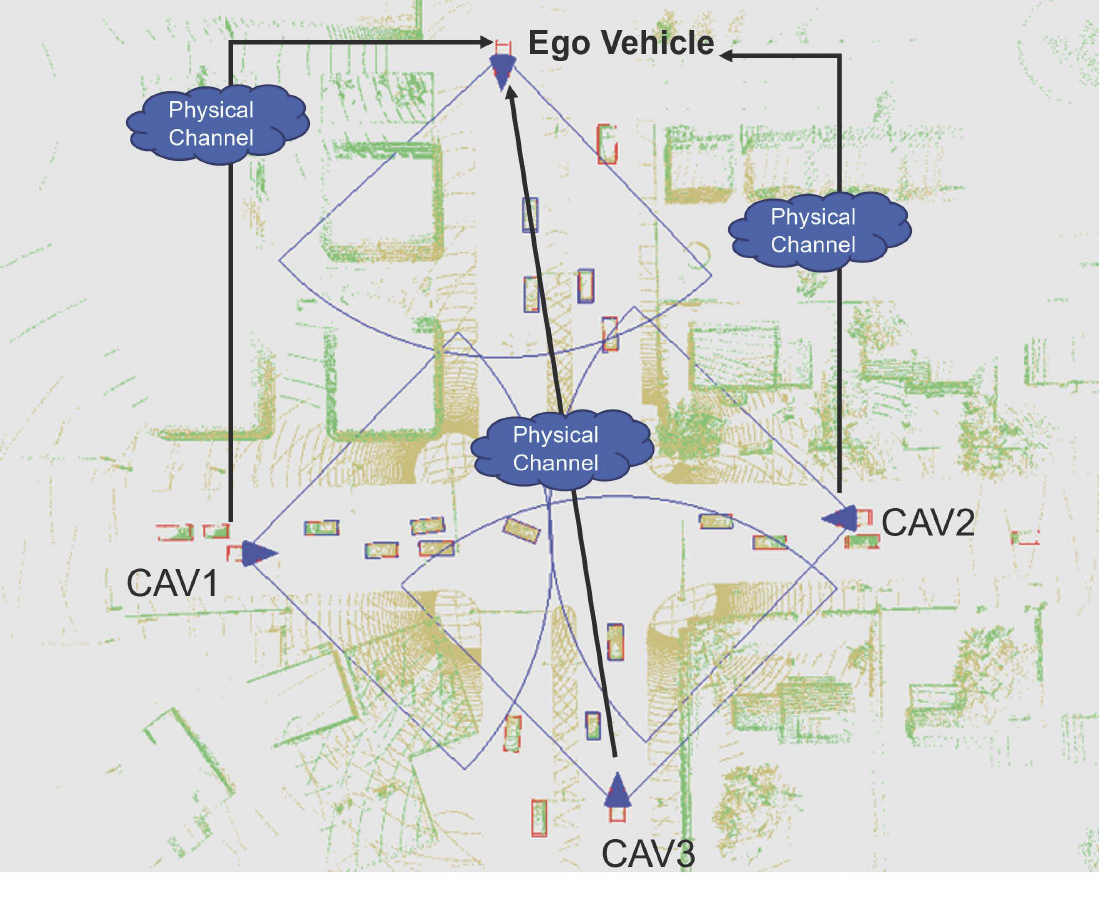}
  \caption{Cooperative perception via V2V communication.}
  \label{v2v_demo}
\end{figure}

\begin{figure*}[!ht]
\centering
\includegraphics[width=0.99\linewidth]{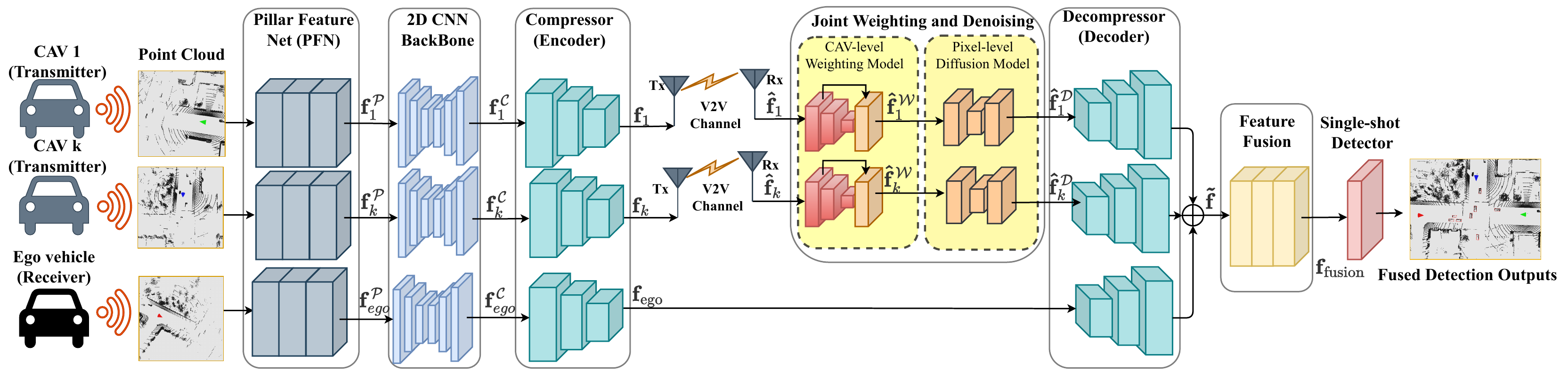}
\caption{System architecture of cooperative perception with V2V communication.}
\label{Fig:architecture}
\end{figure*}

The system model for cooperative perception with V2V communication is illustrated in Fig.~\ref{v2v_demo}. In this model, the ego vehicle collaborates with other CAVs by sharing intermediate features through V2V communications. The fusion process of shared information from multiple CAVs at the ego vehicle can be expressed as 

\begin{equation} 
\mathbf{f}_\text{fusion} = f_{\text{fusion}}(\mathbf{f}_{ego},\mathbf{\hat{f}}_1,\mathbf{\hat{f}}_2,...,\mathbf{\hat{f}}_K ), 
\end{equation} 
where $f(\cdot)$ represents the fusion algorithm employed to combine the shared data received from the CAVs, $\mathbf{f}_{ego}$ denotes the data sensed by the ego vehicle, while $\mathbf{\hat{f}}_k$ refers to the shared information from the $k$-th CAV.
% may be manifested as interference in the cooperative system
The information provided by the CAVs offers diverse viewpoints that enhance both the precision and range of perception. However, when channel conditions degrade significantly, the received information $\mathbf{\hat{f}}_k$ can be distorted due to the communication channel limitations, which in turn compromises the fused feature $\mathbf{f}_\text{fusion}$. 

\subsection{System model for cooperative perception}
    Fig.~\ref{Fig:architecture} shows the system model for cooperative perception with the proposed joint adaptive weighting and denoising in networked vehicular systems. PointPillars \cite{pointpillars} is adopted as the backbone algorithm with a pillar feature net (PFN), a 2D convolutional neural network (CNN) backbone, and a single shot detector (SSD) \cite{ssd}. Firstly, point clouds are divided into an x-y grid of pillars. To address the sparsity of pillars and points, each pillar is augmented with high-dimensional features, where limits are imposed on the number of non-empty pillars and points to formulate a dense tensor. Excess data is randomly sampled, and missing data is zero-padded. Subsequently, a simplified PointNet \cite{PointNet} extracts features with linear layers, batch normalization (BatchNorm), and rectified linear unit (ReLU). The resulting features are scattered back to their original positions, forming a pseudo-image denoted as $\mathbf{f}^\mathcal{P}$ in Fig.~\ref{Fig:architecture}. Then, pseudo-image features ($\mathbf{f}^\mathcal{P}_{ego},\mathbf{f}^\mathcal{P}_1,\dots,\mathbf{f}^\mathcal{P}_k$) are processed by the 2D CNN backbone, which adopts a residual structure with BatchNorm and ReLU layers to produce convoluted features ($\mathbf{f}^\mathcal{C}_{ego},\mathbf{f}^\mathcal{C}_1,\dots,\mathbf{f}^\mathcal{C}_k$). To prepare the sharing features, feature downsampling is conducted at the transmitter using a CNN-based encoder to compress the feature for communication efficiency. At the receiver, the received and convoluted sharing features ($\mathbf{\hat{f}}_1,\dots,\mathbf{\hat{f}}_k$) are concatenated with $\mathbf{f}_{ego}$, respectively, and then deconvoluted to reconstruct the feature map $\mathbf{\tilde{f}}$. This downsampling and upsampling operation could take light channel impairments into account during model training due to its autoencoder (AE) structure. Nevertheless, severe and dynamic channel impairments cannot be addressed by this encoder-decoder structure \cite{guang1}. Therefore, our proposed CAV-level weighting and pixel-level denoising method is implemented at the receiver before up-sampling, which will be introduced in Section \ref{sec3}. Furthermore, feature fusion is conducted to aggregate the concatenated feature $\mathbf{\tilde{f}}$ into $\mathbf{f}_\text{fusion}$. Finally, an SSD is used to output the classification results of objects and the regression results of their box localization.

To enable effective collaboration among multiple CAVs, V2V communications are critical to enable them to share perceptual information. This learning-based detection backbone can be seamlessly integrated with end-to-end communications systems to achieve global optimization. Instead of assuming perfect communication, the detection networks can be trained with light channel impairments to enhance the system robustness. The loss function for the cooperative perception can be expressed as,
\begin{equation}
    \mathcal{L}_{coop} = \frac{1}{N} (\alpha_{reg}\mathcal{L}_{reg} + \alpha_{cls}\mathcal{L}_{cls}) , \label{supervised_loss}
\end{equation}
where $\alpha_{cls}$ and $\alpha_{reg}$ are the parameters for classification and regression loss, $\frac{1}{N}$ denotes the positive number of anchors, $\mathcal{L}_{cls}$ denotes the object classification loss using the focal loss in \cite{8237586}, and $\mathcal{L}_{reg}$ denotes the localization loss that is calculated by localization regression residuals between ground truth and anchors using smooth L1 loss \cite{pointpillars}. This framework serves as the benchmark and our work will focus on further improving cooperative perception under wireless channel impairments.

In this paper, we utilize V2VNet \cite{v2vnet} as an intermediate fusion technique to combine the received features by a graph neural network (GNN). V2VNet is particularly effective in balancing perception performance with the constraints of existing hardware transmission bandwidth by compressing the intermediate representations of the detector. The proposed joint weighting and denoising module will be introduced in Section \ref{sec3}.

\subsection{V2V Channel Model}
To investigate the impact of V2V channel, three communication models are considered: the Rician fading channel with free-space path loss as defined in \eqref{eq1}, the WINNER II channel model \cite{winner2} and a time-varying non-stationary V2V channel as in \eqref{2}. 
Furthermore, to simulate the practical channels, we account for imperfect channel state information (CSI) corrupted by AWGN and use a pilot-based least-squares method for channel estimation, which introduces estimation error. A zero-forcing detector is applied to recover the signal $\mathbf{\hat{f}}_k$ from the received signals. 

\subsubsection{\textbf{Rician fading}} 
This channel considers a single-input single-output (SISO) system with free-space path loss and Rician fading. The information received by the ego vehicle from the $k$-th CAV is denoted as
\begin{equation}
    \mathbf{y}_k = \sqrt{\frac{p_0}{d_k^n}} h_k\mathbf{x}_k + \mathbf{w}_k, \label{eq1}
\end{equation}

\noindent
where $\frac{p_0}{d_k^n}$ denotes the path loss with $p_0$ representing the power loss at 1 meter, $d_k$ denotes the distance between the ego vehicle and the $k$-th CAV, and $n$ is the path loss exponent. The variable $h_k$ characterizes the Rician fading, modeled as a complex normal distribution $\mathcal{CN}(\mu,\sigma_h^2)$ with non-zero mean and parameterized by the Rician factor 
$K= \frac{\mu^2}{\sigma_h^2-\mu^2}$. The vector $\mathbf{y}_k \in \mathbb{C}^{L\times 1}$ represents the complex-valued signals received from the $k$-th CAV and $L$ is the dimension of signals, while $\mathbf{x}_k \in \mathbb{C}^{L\times 1}$ refers to the transmitted signal. Lastly, $\mathbf{w}_k$ denotes the AWGN, following $\mathcal{CN}(0,\sigma^2)$. To incorporate the non-stationarity into this channel model, the Rician factor, $K$, is allowed to vary over time. 

\subsubsection{\textbf{WINNER II}}
The WINNER II model \cite{winner2} is a widely used geometry-based stochastic channel model that captures multi-path propagation effects through large-scale parameters (path loss, shadow fading, delay and angular spreads) and small-scale parameters (delays, powers, and angles of arrival and departure). By representing dynamic multi-path conditions, it can be used for V2V channel modeling. The channel response is obtained by summing the contributions of all multi-path components and can be expressed as
\begin{equation}
    H(t,\tau) =  \sum_{n=1}^{N}\sum_{m=1}^{M} \sqrt{\frac{P_n}{M}}
    e^{j\Phi_{n,m}} \cdot e^{j2\pi v_{n,m} t}\cdot \delta(\tau - \tau_{n,m})
    \label{CIRwinner}
\end{equation}
where $N$ and $M$ denote the number of clusters and the number of rays within each cluster, respectively, $P_n$ is the normalized power of the $n$-th cluster, $\Phi_{n,m}$ represents the random initial phase, $v_{n,m}$ denotes the Doppler frequency, and $\tau_{n,m}$ is the delay associated with the ray.

\subsubsection{\textbf{Non-stationary V2V}}
In V2V communications, the non-stationary wireless channels are affected by the dynamic nature of CAVs, such as their velocity, acceleration, and inter-vehicle spacing. To reflect the temporal non-stationarity, a time-varying V2V channel model \cite{li2020practical} is considered. In this model, CAVs attached with mobile transmitter (Tx) and receiver (Rx) are moving within a traffic scene in trajectories with varying velocities as ${{v}^o}(t)$, where the superscript $o \in \{ {\rm{Tx, Rx}}\}$ represent Tx and Rx in brief. The V2V communication model consists of line-of-sight (LoS) and non-line-of-sight (NLoS) paths. The NLoS paths are described as multiple-bounced propagation paths by applying the twin-cluster model. Under NLoS scenario, propagation paths can be divided into three parts, i.e., Tx to the first cluster $S^{Tx}_{n}$, Rx to the last cluster $S^{Rx}_{n}$, and the rest between $S^{Tx}_{n}$ and $S^{Rx}_{n}$. The $S^{Tx}_{n}$ and $S^{Rx}_{n}$ can be modeled as moving scattering clusters, each characterized by its own movement speed, with the propagation path between the two clusters abstracted as a virtual link \cite{channel1}. Under general V2V traffic scenarios, the V2V channel impulse response (CIR) between the Tx and Rx at time instant $t$ can be expressed as \cite{channel2}
\begin{equation}
    {h}(t,\tau )=\sum\limits_{n=1}^{N}{{{P}_{n}}(t){{{\tilde{h}}}_{n}}(t)}\delta (\tau -{{\tau }_{n}}(t)),
    \label{2}
\end{equation} 
where $N$ indicates the number of propagation paths, ${P}_{n}(t)$ denotes the normalized power of the $n$-th path, ${\tau }_{n}(t)$ represents the delay of the $n$-th path, ${\tilde{h}}_{n}(t)$ can be expressed as
\begin{equation}
    {{\tilde{h}}_{n}}(t)\!=\!\!\frac{1}{\sqrt{M}}\sum\limits_{m=1}^{M}{{}}{{\text{e}}^{\text{j}\cdot (2\pi \cdot \int_{0}^{t}{{}}{{f}_{n,m}}(t')\text{d}t'+{{\Phi }_{n,m}}(t)+{{\theta }_{n,m}})}},
    \label{4}
\end{equation} 
where ${{\Phi }_{n,m}}(t)$ denotes the phase-related parameters induced by the relative position relationship between the scatterers and the Tx and Rx antennas, ${\theta }_{n,m}$ represents the initial random phase and distributed uniformly over $(0,2\pi]$, ${{f}_{n,m}}(t)$ represents the Doppler frequency caused by the movements of CAVs. The parameter ${{f}_{n,m}}(t)$ includes Doppler shifts from both the Tx and Rx, i.e., $f_{n,m}(t) = f_{n,m}^{Tx}(t) + f_{n,m}^{Rx}(t)$. Here, $f_{n,m}^o(t)$, where $o\in\{Tx,Rx\}$, can be calculated as
\begin{equation}
f_{n,m}^o(t) = \;\frac{{{v^o}(t)}}{\lambda } \cdot \cos (\alpha _{n,m}^o(t) - \alpha _v^o(t)),
\end{equation}\\
where $\alpha_v^o$ represents the time-variant moving direction of Tx or Rx, and $\alpha _{n,m}^{o}(t)$ denotes the angle of arrival (AoA) or angle of departure (AoD). Furthermore, introducing acceleration into the model enables simulation of time-varying Tx and Rx speeds, leading to a more realistic characterization of the actual channel. The detailed calculation process of the above channel parameters can be found in \cite{li2020practical}. 

By incorporating vehicles dynamic characteristics, like poses, velocities, and accelerations into ${{\tilde{h}}_{n}}(t)$, the non-stationary V2V channel model is practical to reveal the impact of vehicle-related variations on the non-stationary channel characteristics. 

\section{Joint adaptive weighting and denoising approach}\label{sec3}
\begin{figure}[!htp]
    \centering   \includegraphics[width=0.48\textwidth]{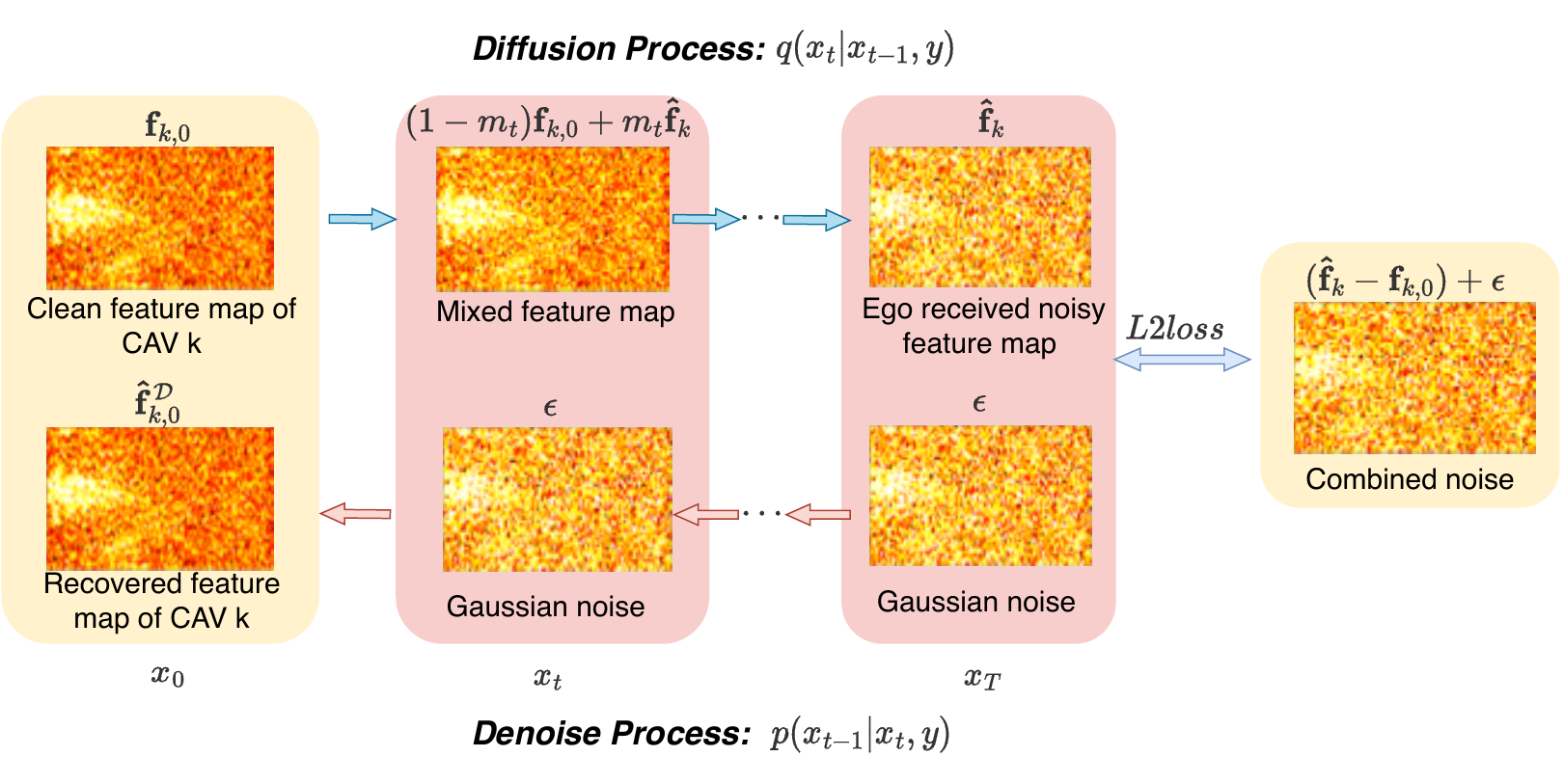}
    \caption{Pixel-level diffusion and denoising processes for feature maps in cooperative perception system.}
    \label{Fig.diff}
\end{figure}
In this section, a hierarchical joint feature processing framework is presented, which combines CAV-level self-supervised weighting and pixel-level denoising based on generative learning. The synergy between these two modules is designed to overcome the limitations of standalone CAV-level or pixel-level processing. The proposed architecture integrates weighting and denoising to leverage the simplicity of CAV weighting while enabling fine-grained restoration through generative diffusion models. By incorporating contrastive self-supervision with generative learning, the framework achieves robust feature representation and synthesis, effectively addressing the non-stationarity of V2V channel impairments. In addition, the shared use of residual blocks and skip connections ensures structural compatibility between the two modules, enabling seamless adaptation and efficient integration.

\subsection{Diffusion for pixel-level denoising}
In V2V communication, cooperative perception performance is affected by communication channel impairments, such as noise, fading and time-varying disturbances, that may corrupt intermediate feature maps. Existing methods mainly rely on supervised distortion-in-the-loop training to address this, which lacks robustness against severe distortions. Inspired by the successful application of diffusion model in voice signals \cite{cdiffse}, we propose a pixel-level denoising module based on the conditional diffusion process to address the distorted shared feature maps. 

The conditional diffusion model consists of diffusion process and reverse denoising process. The diffusion process estimates the mixed noise between the transmitted clean feature $x_0$ and received distorted feature $y$. The reverse process generates denoised feature maps. 

The conditional diffusion process can be expressed as,
\begin{gather}
  q_{\text{cond}}(x_t|x_0,y) = \mathcal{N}(x_t;(1-m_t)\sqrt{\bar{\alpha}_t}x_0 + m_t\sqrt{\bar{\alpha}_t}y ,\delta_tI), \label{diff:eq4}\\
  \bar{\alpha}_t =  \prod_{i=1}^{t} \alpha_i,\\
  m_t=\sqrt{(1 - \bar{\alpha}_t)/\sqrt{\bar{\alpha}_t}}, \\
  \delta_t = (1-\bar{\alpha}_t)-m_t^2\bar{\alpha}_t,
  \label{diff:eq6}
\end{gather}

\noindent
where $\beta_t$ denotes the predefined noise schedule at diffusion step $t$ and $\alpha_t = 1-\beta_t$. The mean of $x_t$ is formulated as a step-dependent linear interpolation between the clean feature $x_0$ and the distorted feature $y$, controlled by the coefficient $m_t$. The interpolation coefficient satisfies $m_0 = 0$ and increases to approximately $m_T \approx 1$, thereby gradually shifting the diffusion state from being centered on $x_0$ to being centered on $y$. The variance term $\delta_t$ determines the stochastic perturbation introduced at each diffusion step. The linear interpolation parametrizes a smooth transition between clean and distorted feature representations in feature space. By controlling the contribution of $x_0$ and $y$ through $m_t$ and introducing Gaussian perturbation, the model learns the conditional distribution of distorted features around the clean representation.

The conditional reverse process is to estimate the posterior distribution of the latent variable $x_{t-1}$, given the current diffusion state $x_t$ and the received distorted feature $y$. Unlike traditional diffusion models, the conditional reverse process initializes the reverse process from the terminal conditional diffusion state
\begin{gather}
     p_{\text{cond}}(x_{T}|y) = \mathcal{N}(x_{T};\sqrt{\bar{\alpha}_T}y,\tilde{\delta}_TI),
\end{gather}
which corresponds to the final state of the forward conditional diffusion process when $m_T \approx 1$. At each reverse step, the conditional reverse process is parameterized as,
\begin{gather}
  p_{\text{cond}}(x_{t-1}|x_t,y) = \mathcal{N}(x_{t-1};\mu_\theta(x_t,y,t),\tilde{\delta}_tI),
  \label{eq:new reverse}
\end{gather}
where $\mu_\theta(x_t,y,t)$ is the learned mean and $\tilde{\delta}_t$ is the reverse variance.

Since the forward diffusion state $x_t$ is constructed as a linear interpolation between $x_0$ and $y$ with additive noise, the reverse mean is modeled as a linear combination of the current state $x_t$, the conditioning feature $y$, and a predicted noise component:
\begin{gather}
     \mu_\theta(x_t,y,t) = c_{xt}x_{t} + c_{yt} y - c_{\epsilon t} \epsilon_\theta(x_t,y,t)\label{eq:new reverse mean}.
\end{gather}
\begin{gather}
    c_{xt} = \frac{1-m_t}{1-m_{t-1}}\frac{\delta_{t-1}}{\delta_{t}}\sqrt{\alpha}_t + (1-m_{t-1})\frac{\delta_{t|t-1}}{\delta_{t}}\frac{1}{\sqrt{\alpha}_t}, \label{eq:c_{xt}} \\
    c_{yt} = (m_{t-1}\delta_t - \frac{m_t(1-m_t)}{1-m_{t-1}}\alpha_t\delta_{t-1})\frac{\sqrt{\bar{\alpha}_{t-1}}}{\delta_t}, \label{eq:c_{yt}} \\
    c_{\epsilon t} = (1-m_{t-1})\frac{\delta_{t|t-1}}{\delta_t}\frac{\sqrt{1-\bar{\alpha}_t}}{\sqrt{\alpha_t}},
  \label{eq:c_epsilon}
\end{gather}
where the conditional one-step variance of the forward diffusion process is defined as
\begin{equation}
    \delta_{t|t-1} = \delta_t - \left(\frac{1-m_t}{1-m_{t-1}}\right)^2 \alpha_t \delta_{t-1}.
  \label{diff:eq15}
\end{equation}
The variance in the reverse process is defined as,
\begin{equation}
    \tilde{\delta}_t = \frac{\delta_{t|t-1} * \delta_t}{\delta_{t-1}},
  \label{diff:eq16}
\end{equation}
which matches the posterior variance derived from the conditional diffusion formulation. Under this formulation, the reverse process iteratively refines $x_t$ by removing the predicted composite noise component while incorporating guidance from the received distorted feature $y$. After $T$ reverse steps, the denoised feature representation $x_0$ is recovered.

The learnable parameters $\theta$ are optimized through the generative learning process. The evidence lower bound (ELBO) optimization \cite{ddpm} for the conditional diffusion process can be expressed as,
\begin{equation}
\begin{split}
    ELBO = c^\prime + \sum_{t=1}^{T}\kappa^\prime_t \mathbb{E}_{x_0, \epsilon, y} \left\| \frac{m_t\sqrt{\bar{\alpha}_t}}{\sqrt{1 - \bar{\alpha}_t}}{(y - x_0)} \right. \\
    \left. + \frac{\sqrt{\delta_t}}{\sqrt{1 - \bar{\alpha}_t}}\epsilon - \epsilon_\theta(x_t, y, t) \right\|^2_2,
\end{split}
  \label{eq:ELBO_optimized}
\end{equation}
where $c^\prime$ and $\kappa^\prime_t$ denote constant parameters, and $\epsilon$ denotes the Gaussian noise in $x_t$. The first term inside the norm corresponds to the structured distortion component $(y-x_0)$, while the second term represents stochastic Gaussian noise. Accordingly, the network $\epsilon_\theta(x_t,y,t)$ is trained to jointly estimate both components by minimizing the L2 loss between the predicted noise and the true mixed noise in $x_t$. The corresponding gradient update is given by,
\begin{equation}
\scalebox{0.9}{
    $\nabla_\theta \left\| \frac{1}{\sqrt{1 - \bar{\alpha}_t}} 
    \left( m_t \sqrt{\bar{\alpha}_t} (y - x_0) + \sqrt{\delta_t} \epsilon \right)
    - \epsilon_\theta(x_t, y, t) \right\|^2_2$.}
 \label{eq:gradient}
\end{equation}
Finally, $x_0$ can be recovered by a $T$-step iterative reverse process.

To apply the conditional diffusion model to cooperative perception, Fig.~\ref{Fig.diff} illustrates the forward diffusion and reverse denoising processes operating on shared feature maps. The diffusion variables are mapped to perception features as follows: the clean variable $x_0$ corresponds to the transmitted intermediate feature $\mathbf{f}_{k,0}$ of the $k$-th CAV; the conditioning variable $y$ corresponds to the received distorted feature $\mathbf{\hat{f}}_k$ at the ego vehicle; and the latent diffusion state $x_t$ corresponds to the feature $\mathbf{\hat{f}}^{\mathcal{D}}_{k,t}$ at diffusion step $t$, where the superscript $\mathcal{D}$ denotes features evolving within the denoising process.

In the forward diffusion process, the clean feature map $\mathbf{f}_{k,0}$ is progressively corrupted over $T$ steps by adding Gaussian noise while being conditioned on the received feature $\mathbf{\hat{f}}_k$. This produces a sequence of diffused feature maps $\{\mathbf{\hat{f}}^{\mathcal{D}}_{k,t}\}_{t=1}^{T}$ sampled from the distribution defined in Eq.~\eqref{diff:eq4}. The terminal state $x_T$ corresponds to a noisy feature map centered around the received feature $\mathbf{\hat{f}}_k$, capturing both channel-induced distortion and stochastic diffusion noise under dynamic V2V channel conditions. This multi-step diffusion procedure does not model the physical evolution of the V2V communication channel. Instead, it serves as a conditional generative modeling mechanism that learns the statistical relationship between clean and distorted feature representations.

In the reverse denoising process, the iteration is initialized from $\mathbf{\hat{f}}^{\mathcal{D}}_{k,T}$ and progressively refines the latent feature by predicting and removing the combined noise at each step, conditioned on $\mathbf{\hat{f}}_k$, according to Eq.~\eqref{eq:new reverse}. After $T$ reverse steps, the denoised feature map $\mathbf{\hat{f}}^{\mathcal{D}}_{k,0}$ is obtained and subsequently used for cooperative feature fusion.

\subsection{Joint weighting and denoising}
\begin{figure}[htb]
\centering
\includegraphics[width=0.90\linewidth]{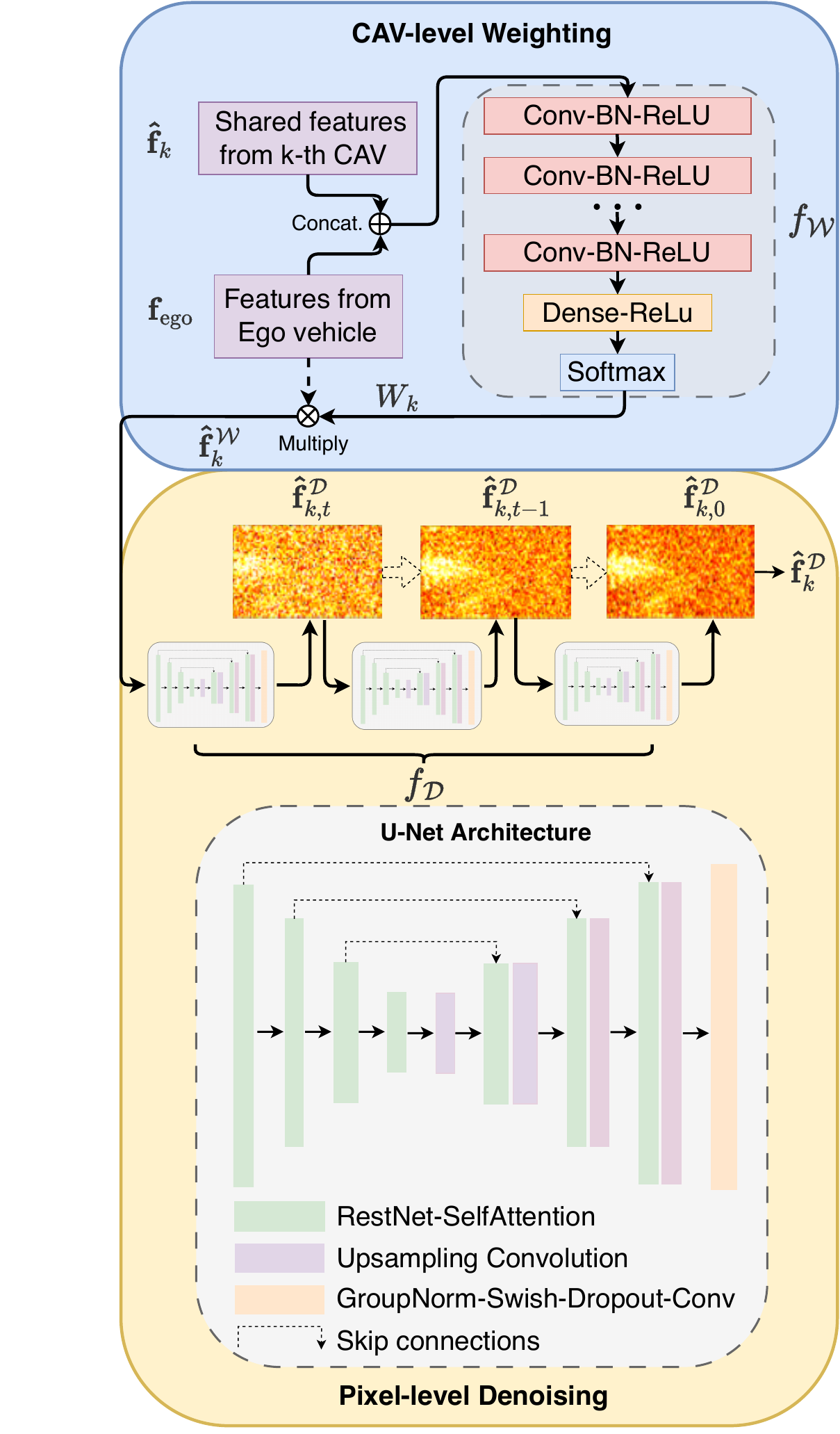}
\caption{The proposed joint weighting and denoising algorithm pipeline.}
\label{Fig:joint}
\end{figure}

As shown in Fig. \ref{Fig:joint}, the proposed joint CAV-level weighting and pixel-level denoising aims to offer an effective approach for fine-grained feature enhancements in cooperative perception. The weighting module can be expressed as,
\begin{gather}
    W_k = f_{\mathcal{W}}(\mathbf{f}_{\text{ego}}, \mathbf{\hat{f}}_k) \label{weight1}, \\ 
    \mathbf{\hat{f}}_{k}^{\mathcal{W}} = W_k \cdot \mathbf{\hat{f}}_k, \label{weight2} 
\end{gather}

\noindent
where $W_k$ denotes the weighting output with a value from 0 to 1, $f_{\mathcal{W}}$ denotes the weighting module, $\mathbf{f}_{\text{ego}}$ and $\mathbf{\hat{f}}_k$ denote the feature from the ego vehicle and the received feature from the $k$-th CAV, respectively, and $\mathbf{\hat{f}}_{k}^{\mathcal{W}}$ is the $k$-th weighted feature. This weighting module is firstly proposed in \cite{guang2}, which adopts self-supervised training with contrastive features of shared features and the ego feature to mitigate the adverse effects of channel distortion at the vehicle level. One key reason for adopting this module in the joint framework is the efficiency and adaptivity of using one value to filter out severe distorted features. However, when there is moderate distortion, it may also amplify noise, making it harder for the decoder to recover the original information.

To solve this, a U-Net architecture is proposed as the denoising network based on diffusion process to enhance features with moderate distortion. As shown in Fig. \ref{Fig:joint}, the output features from the weighting module are iteratively processed by the denoising module, which can be expressed as,
\begin{equation}
    \mathbf{\hat{f}}_{k}^{\mathcal{D}} = f_{\mathcal{D}}(\mathbf{\hat{f}}_{k}^{\mathcal{W}}),\label{denoise}
\end{equation}

\noindent
where $\mathbf{\hat{f}}_{k}^{\mathcal{D}}$ is the denoised features and $f_{\mathcal{D}}$ denotes the denoising function based on the U-Net architecture \cite{diff_sr3}.

The U-Net architecture incorporates multiple network blocks, including ResNet-SelfAttention, upsampling convolution, and a composite block consisting of GroupNorm (Group Normalization), Swish activation, dropout, and convolution, interconnected through direct and skip connections. Specifically, the ResNet-SelfAttention components are designed to effectively capture local features while simultaneously integrating global contextual information, leveraging long-range dependencies by the attention mechanism. The upsampling convolution layers progressively increase the spatial resolution of feature maps, facilitating the reconstruction of data at finer resolutions during the generative process. Skip connections link these upsampled features with features from earlier layers, aiding in the recovery of fine details. Finally, the integration of group normalization, Swish activation, dropout, and convolution provides robust handling of noisy inputs at varying levels while ensuring feature consistency across iterative denoising steps. 

The above learnable parameters are optimized by maximizing the ELBO in Eq.~\eqref{eq:ELBO_optimized}, which reduces to minimizing the L2 loss between the predicted noise $\epsilon_\theta(x_t,y,t)$ and the ground-truth combined noise in the diffusion state. In the cooperative perception system, the clean feature $x_0$ corresponds to the transmitted intermediate feature $\mathbf{f}_{k,0}$, while in the joint weighting and denoising framework the conditioning variable $y$ corresponds to the weighted received feature $\mathbf{\hat{f}}^{\mathcal{W}}_k$. The latent diffusion state $x_t$ corresponds to $\mathbf{\hat{f}}^{\mathcal{D}}_{k,t}$. The forward process progressively corrupts $\mathbf{f}_{k,0}$ toward $\mathbf{\hat{f}}^{\mathcal{W}}_k$ over $T$ steps, producing a stochastic terminal state $\mathbf{\hat{f}}^{\mathcal{D}}_{k,T}$ centered around the conditioning feature. The reverse denoising process is initialized from this terminal state and iteratively removes the predicted composite noise conditioned on $\mathbf{\hat{f}}^{\mathcal{W}}_k$. After $T$ reverse iterations, the recovered feature $\mathbf{\hat{f}}^{\mathcal{D}}_{k,0}$ serves as the denoised representation for cooperative feature fusion.

To integrate the training of diffusion model with cooperative perception, the loss function can be expressed by, 
\begin{equation}
    \mathcal{L}_\text{denoise} =\beta_\text{coop}\mathcal{L}_\text{coop} + \beta_\text{diffusion}\mathcal{L}_\text{diffusion},  \label{diffusion_loss}
\end{equation}
where $\mathcal{L}_\text{coop}$ denotes the object detection loss in \eqref{supervised_loss}, and $\mathcal{L}_\text{diffusion}$ denotes the mean square error loss between the predicted and true noise in the diffusion process. 

\section{Numerical Results and Discussion}\label{sec4}
In this section, we evaluate the performance of our proposed joint weighting and denoising approach for cooperative perception in networked vehicular systems with various V2V communication channels.  

\subsection{Simulation Setup}
The simulation settings are as below:

\subsubsection{Dataset}
To simulate real-world cooperative perception with V2V communication, we exploit the V2V4Real dataset \cite{v2v4real}, a large-scale, multi-modal dataset collected from real-world vehicles and traffic scenarios for cooperative perception. This dataset encompasses a variety of driving environments, including intersections, highway entrance ramps, straight highway segments, and straight city roads, covering a total distance of 410 kilometers. It includes 20$K$ LiDAR frames and 240$K$ annotated 3D bounding boxes, providing a comprehensive foundation for both training and evaluation. \hlbreakable{Additionally, the OPV2V dataset\cite{openv2v} is used to evaluate the proposed models in multi-CAV scenarios, while V2XSet\cite{v2x-vit} is used to assess generalization across different fusion methods and cooperative perception frameworks.}

\subsubsection{Baseline}
For the cooperative perception, PointPillars \cite{pointpillars} with V2VNet \cite{v2vnet} fusion module is used as the backbone for cooperative perception. The benchmark models are as below:
\begin{itemize}
    \item Single AV detection is adopted by using only the ego vehicle, which does not suffer from distorted shared information, but also could not benefit from other cooperation.
    \item Cooperative perception with CNN-based AE (\textit{Coop}) is adopted, which is widely used in cooperative perception baselines \cite{openv2v,v2x-vit,cobevt,v2v4real}. Additionally, we apply the supervised distortion-in-the-loop training strategy \cite{guang1} to simulate a certain degree of communication signal distortion to improve the model robustness as benchmark model.
    \item Cooperative perceptions with only CAV-level weighting (\textit{Coop-W}) \cite{guang2} is used for the ablation study of the proposed joint weighting and denoising approach. 
    \item Cooperative perception with pixel-level denoising (\textit{Coop-D}) is adopted to benchmark the effectiveness of the proposed denoising module.
\end{itemize}
\hlbreakable{Additionally, we integrate the proposed weighting and denoising modules with attentive fusion~\cite{openv2v}, Where2comm~\cite{where2comm}, and V2X-ViT~\cite{v2x-vit} to validate their generalization across different fusion approaches.} A compression ratio of 32 is adopted in the encoder–decoder module to reduce communication overhead. Under this commonly used setting, feature compression has limited impact on detection performance and does not affect the relative improvement introduced by the proposed weighting and diffusion modules. Average precision (AP) is utilized as the performance metric, which calculates the AP at different levels of thresholds in terms of the Intersection-over-Union (IoU).

\subsubsection{Communication Settings}
To validate the effectiveness of our model in different communication conditions, the non-stationary V2V channel is adopted for the model training with a random SNR from 15 to 20 dB. The WINNER II \cite{winner2} and the Rician fading channel are used for testing as the realistic and simulated channel model, respectively. Additionally, time-varying noise levels and CSI errors are considered for validation, where the variations are simulated as a Gaussian process following $\mathcal{N}(0,\sigma^2)$.

\subsubsection{Training}

Firstly, the adaptive weighting module is trained by the self-supervised learning scheme in \cite{guang2} without fine-tuning with the detection backbone. Subsequently, the diffusion model is trained with the loss function in \eqref{diffusion_loss} with $\beta_\text{coop}=0.1$ and $\beta_\text{diffusion}=1$. We did not perform end-to-end fine-tuning to fairly evaluate the contribution of the diffusion module without the influence of joint optimization of the entire network. In addition, keeping the backbone fixed allows the diffusion module to be used as a plug-in component in existing cooperative perception systems without requiring full retraining. For fair comparison, all variants including Coop, Coop-W, Coop-D, and Coop-WD share the same cooperative perception backbone, training data splits, and non-stationary V2V channel settings. The weighting and diffusion modules are trained under identical noise distributions, and the backbone remains frozen when training the denoising module, ensuring that performance differences arise solely from the inclusion of the weighting and diffusion components.

The U-Net architecture is adopted as the diffusion model which contains three residual blocks and four depth multipliers of $\{1,2,4,8\}$. The training noise schedule is linearly spaced as $\beta_t \in [1 \times 10^{-4}, 0.035]$ with 50 diffusion steps. Besides, the fast sample scheme \cite{cdiffse} is used in the reverse process with the inference schedule $[0.0001, 0.001, 0.01, 0.05, 0.2, 0.35]$.

The models are trained and evaluated by Intel(R) Core(TM) i9-14900K CPU and NVIDIA GeForce RTX 4090 GPU.

\begin{table*}[!ht]
\caption{Average precision under various channels. \textbf{Bold} numbers indicate the highest precision and the second-highest is \underline{underlined}.}
\label{table1}\renewcommand{\arraystretch}{0.9}
\setlength{\tabcolsep}{4pt}
\begin{tabular}{c|cccccc|cccccc|cccccc}
\toprule
 &
  \multicolumn{6}{c|}{WINNER II channel} &
  \multicolumn{6}{c|}{Non-stationary V2V channel} &
  \multicolumn{6}{c}{Rician fading} 

\\ \midrule
 SNR &
  \multicolumn{2}{c|}{0 dB} &
  \multicolumn{2}{c|}{10 dB} &
  \multicolumn{2}{c|}{20 dB} &
  \multicolumn{2}{c|}{0 dB} &
  \multicolumn{2}{c|}{10 dB} &
  \multicolumn{2}{c|}{20 dB} &
  \multicolumn{2}{c|}{0 dB} &
  \multicolumn{2}{c|}{10 dB} &
  \multicolumn{2}{c}{20 dB} 
   \\ \midrule
 IoU &
  \multicolumn{1}{c}{@0.5} &
  \multicolumn{1}{c|}{@0.7} &
  \multicolumn{1}{c}{@0.5} &
  \multicolumn{1}{c|}{@0.7} &
  \multicolumn{1}{c}{@0.5} &
  @0.7 &
  \multicolumn{1}{c}{@0.5} &
  \multicolumn{1}{c|}{@0.7} &
  \multicolumn{1}{c}{@0.5} &
  \multicolumn{1}{c|}{@0.7} &
  \multicolumn{1}{c}{@0.5} &
  @0.7 &
    \multicolumn{1}{c}{@0.5} &
  \multicolumn{1}{c|}{@0.7} &
  \multicolumn{1}{c}{@0.5} &
  \multicolumn{1}{c|}{@0.7} &
  \multicolumn{1}{c}{@0.5} &
   \multicolumn{1}{c}{@0.7} 
   \\ \midrule
Single AV &
  \multicolumn{1}{c}{0.398} &
  \multicolumn{1}{c|}{0.220} &
  \multicolumn{1}{c}{0.398} &
  \multicolumn{1}{c|}{0.220} &
  \multicolumn{1}{c}{0.398} &
  \multicolumn{1}{c|}{0.220} &
  
  \multicolumn{1}{c}{0.398} &
  \multicolumn{1}{c|}{0.220} &
  \multicolumn{1}{c}{0.398} &
  \multicolumn{1}{c|}{0.220} &
  \multicolumn{1}{c}{0.398} &
  \multicolumn{1}{c|}{0.220} &
  
  \multicolumn{1}{c}{0.398} &
  \multicolumn{1}{c|}{0.220} &
  \multicolumn{1}{c}{0.398} &
  \multicolumn{1}{c|}{0.220} &
  \multicolumn{1}{c}{0.398} &
  \multicolumn{1}{c}{0.220} 
   \\ \midrule
Coop &
  \multicolumn{1}{c}{0.095} &
  \multicolumn{1}{c|}{0.036} &
  \multicolumn{1}{c}{0.336} &
  \multicolumn{1}{c|}{0.129} &
  \multicolumn{1}{c}{0.479} &
  \multicolumn{1}{c|}{0.206} &
  
  \multicolumn{1}{c}{0.024} &
  \multicolumn{1}{c|}{0.008} &
  \multicolumn{1}{c}{0.103} &
  \multicolumn{1}{c|}{0.046} &
  \multicolumn{1}{c}{0.398} &
  \multicolumn{1}{c|}{0.209} &
  
  \multicolumn{1}{c}{0.141} &
  \multicolumn{1}{c|}{0.042} &
  \multicolumn{1}{c}{0.387} &
  \multicolumn{1}{c|}{0.152} &
  \multicolumn{1}{c}{0.494} &
  \multicolumn{1}{c}{0.232} 
   \\ \midrule
   Coop-W &
  \multicolumn{1}{c}{\underline{0.447}} &
  \multicolumn{1}{c|}{\underline{0.242}} &
  \multicolumn{1}{c}{0.492} &
  \multicolumn{1}{c|}{0.237} &
  \multicolumn{1}{c}{0.545} &
  \multicolumn{1}{c|}{0.222} &
  
  \multicolumn{1}{c}{\underline{0.483}} &
  \multicolumn{1}{c|}{\textbf{0.273}} &
  \multicolumn{1}{c}{\underline{0.522}} &
  \multicolumn{1}{c|}{\underline{0.273}} &
  \multicolumn{1}{c}{0.553} &
  \multicolumn{1}{c|}{0.232} &
  
  \multicolumn{1}{c}{\underline{0.481}} &
  \multicolumn{1}{c|}{\textbf{0.280}} &
  \multicolumn{1}{c}{0.470} &
  \multicolumn{1}{c|}{0.271} &
  \multicolumn{1}{c}{0.461} &
  \multicolumn{1}{c}{0.265} 
   \\ \midrule
Coop-D &
  \multicolumn{1}{c}{0.354} &
  \multicolumn{1}{c|}{0.181} &
  \multicolumn{1}{c}{\underline{0.541}} &
  \multicolumn{1}{c|}{\underline{0.273}} &
  \multicolumn{1}{c}{\underline{0.603}} &
  \multicolumn{1}{c|}{\underline{0.310}} &
  
  \multicolumn{1}{c}{0.151} &
  \multicolumn{1}{c|}{0.072} &
  \multicolumn{1}{c}{0.362} &
  \multicolumn{1}{c|}{0.182} &
  \multicolumn{1}{c}{\underline{0.554}} &
  \multicolumn{1}{c|}{\underline{0.289}} &
  
  \multicolumn{1}{c}{0.345} &
  \multicolumn{1}{c|}{0.190} &
  \multicolumn{1}{c}{\underline{0.477}} &
  \multicolumn{1}{c|}{\underline{0.272}} &
  \multicolumn{1}{c}{\underline{0.513}} &
  \multicolumn{1}{c}{\textbf{0.292}} 
   \\ \midrule
Coop-WD &
  \multicolumn{1}{c}{\textbf{0.462}} &
  \multicolumn{1}{c|}{\textbf{0.266}} &
  \multicolumn{1}{c}{\textbf{0.586}} &
  \multicolumn{1}{c|}{\textbf{0.305}} &
  \multicolumn{1}{c}{\textbf{0.623}} &
  \multicolumn{1}{c|}{\textbf{0.330}} &
  
  \multicolumn{1}{c}{\textbf{0.487}} &
  \multicolumn{1}{c|}{\underline{0.260}} &
  \multicolumn{1}{c}{\textbf{0.543}} &
  \multicolumn{1}{c|}{\textbf{0.288}} &
  \multicolumn{1}{c}{\textbf{0.617}} &
  \multicolumn{1}{c|}{\textbf{0.309}} &
   
   \multicolumn{1}{c}{\textbf{0.483}} &
  \multicolumn{1}{c|}{\underline{0.264}} &
  \multicolumn{1}{c}{\textbf{0.534}} &
  \multicolumn{1}{c|}{\textbf{0.285}} &
  \multicolumn{1}{c}{\textbf{0.549}} &
  \multicolumn{1}{c}{\underline{0.286}} 
   \\ 
   \bottomrule
\end{tabular}
\end{table*}

\subsection{Performance in various wireless channels}
\begin{table}[!ht]
\caption{Average precision with point pillar and attentive fusion as benchmark in Rician fading channel.}
\label{table2}\renewcommand{\arraystretch}{0.9}
\setlength{\tabcolsep}{2.5pt}
\begin{tabular}{c|cc|cc|cc|cc}
\toprule

  \multicolumn{9}{c}{Point pillars with attentive fusion} 

\\ \midrule
 SNR &
  \multicolumn{2}{c|}{0 dB} &
  \multicolumn{2}{c|}{10 dB} &
  \multicolumn{2}{c|}{20 dB}  &
  \multicolumn{2}{c}{30 dB} 
   \\ \midrule
 IoU &
  \multicolumn{1}{c}{@0.5} &
  \multicolumn{1}{c|}{@0.7} &
  \multicolumn{1}{c}{@0.5} &
  \multicolumn{1}{c|}{@0.7} &
  \multicolumn{1}{c}{@0.5} &
  \multicolumn{1}{c|}{@0.7} &
  \multicolumn{1}{c}{@0.5} &
  \multicolumn{1}{c}{@0.7} 
   \\ \midrule
CNN-based AE &
  \multicolumn{1}{c}{0.295} &
  \multicolumn{1}{c|}{0.177} &
  \multicolumn{1}{c}{0.442} &
  \multicolumn{1}{c|}{0.272} &
  \multicolumn{1}{c}{0.463} &
  \multicolumn{1}{c|}{0.288}  &
  \multicolumn{1}{c}{0.467} &
  \multicolumn{1}{c}{0.291} 
   \\ \midrule
   Weighting only&
  \multicolumn{1}{c}{\underline{0.461}} &
  \multicolumn{1}{c|}{\textbf{0.275}} &
  \multicolumn{1}{c}{0.465} &
  \multicolumn{1}{c|}{\textbf{0.289}} &
  \multicolumn{1}{c}{0.466} &
  \multicolumn{1}{c|}{0.291} &
   \multicolumn{1}{c}{0.467} &
  \multicolumn{1}{c}{0.291} 
   \\ \midrule
Denoising only&
  \multicolumn{1}{c}{0.451} &
  \multicolumn{1}{c|}{0.266} &
  \multicolumn{1}{c}{\underline{0.526}} &
  \multicolumn{1}{c|}{0.280} &
  \multicolumn{1}{c}{\underline{0.566}} &
  \multicolumn{1}{c|}{\underline{0.302}}&
  \multicolumn{1}{c}{\textbf{0.578}} &
  \multicolumn{1}{c}{\textbf{0.306}} 
   \\ \midrule
Weighting \& &
  \multicolumn{1}{c}{\multirow{2}{*}{\textbf{0.468}}} &
  \multicolumn{1}{c|}{\multirow{2}{*}{\underline{0.270}}} &
  \multicolumn{1}{c}{\multirow{2}{*}{\textbf{0.531}}} &
  \multicolumn{1}{c|}{\multirow{2}{*}{\underline{0.283}}} &
  \multicolumn{1}{c}{\multirow{2}{*}{\textbf{0.569}}} &
  \multicolumn{1}{c|}{\multirow{2}{*}{\textbf{0.303}}} &
  \multicolumn{1}{c}{\multirow{2}{*}{\underline{0.576}}} &
  \multicolumn{1}{c}{\multirow{2}{*}{\textbf{0.306}}} \\ denoising& & & & & & & &
   \\ 
   \bottomrule
\end{tabular}
\end{table}

\begin{table}[!ht]
\caption{\hlbreakable{Comparison of average precision after integrating the proposed weighting and denoising module with V2X-ViT~\cite{v2x-vit} and Where2comm~\cite{where2comm} on the V2XSet dataset~\cite{v2x-vit}.}}
\centering
\label{table_v2x}\renewcommand{\arraystretch}{0.9}
\setlength{\tabcolsep}{2.5pt}
\begin{tabular}{c|cccccc}
\toprule
   \multicolumn{7}{c}{Non-stationary V2V channel} 
\\
\toprule
  SNR&
  \multicolumn{2}{c|}{0 dB} &
  \multicolumn{2}{c|}{10 dB} &  
  \multicolumn{2}{c}{20 dB} 

   \\ \midrule
 IoU&
  \multicolumn{1}{c}{ @0.5 } &
  \multicolumn{1}{c|}{ @0.7 } &
  \multicolumn{1}{c}{ @0.5 } &
  \multicolumn{1}{c|}{ @0.7 } &
  \multicolumn{1}{c}{ @0.5 } &
  \multicolumn{1}{c}{ @0.7 } 

  \\ \toprule
    V2X-ViT&
  \multicolumn{1}{c}{0.325} &
  \multicolumn{1}{c|}{0.205} &
  \multicolumn{1}{c}{0.493} &
  \multicolumn{1}{c|}{0.320} &
  \multicolumn{1}{c}{0.664} &
  \multicolumn{1}{c}{0.462} \\
  \midrule

  V2X-ViT + Coop-WD &
  \multicolumn{1}{c}{\underline{0.605}} &
  \multicolumn{1}{c|}{\textbf{0.497}} &
  \multicolumn{1}{c}{\textbf{0.705}} &
  \multicolumn{1}{c|}{\textbf{0.565}} &
  \multicolumn{1}{c}{\textbf{0.787}} &
  \multicolumn{1}{c}{\textbf{0.581}} \\

 \toprule

  Where2comm &
  \multicolumn{1}{c}{0.029} &
  \multicolumn{1}{c|}{0.008} &
  \multicolumn{1}{c}{0.312} &
  \multicolumn{1}{c|}{0.141} &
  \multicolumn{1}{c}{0.728} &
  \multicolumn{1}{c}{0.455} \\

   \midrule

  Where2comm + Coop-WD&
  \multicolumn{1}{c}{\textbf{0.691}} &
  \multicolumn{1}{c|}{\underline{0.476}} &
  \multicolumn{1}{c}{\underline{0.693}} &
  \multicolumn{1}{c|}{\underline{0.477}} &
  \multicolumn{1}{c}{\underline{0.744}} &
  \multicolumn{1}{c}{\underline{0.490}} 

    \\ \bottomrule
\end{tabular}
\end{table}
In this section, we evaluate the cooperative perception under three different channels: Rician fading, WINNER II, and non-stationary V2V channel. The model is trained with the non-stationary V2V channel and tested on the Rician fading, WINNER II channel and the non-stationary V2V channel to validate its generalizability on unseen channels.

As shown in Table~\ref{table1}, the cooperative perception with the proposed diffusion model has the best robustness among all evaluated models when SNR ranges from 0 to 20 dB. Additionally, it achieves the highest accuracy as channel conditions improve. In comparison, single CAV perception, which relies solely on the ego vehicle's own sensing data to avoid distortions from shared information among CAVs, achieves consistently low AP scores: $39.8\%$ at IoU=0.5, and $22\%$ at IoU=0.7, regardless of the SNR. \textit{Coop-W}, which employs adaptive weighting to filter out distorted information, effectively mitigates the impact of severe channel impairments. However, it experiences slight performance degradation at IoU=0.7 when the SNR increase from 10 to 20 dB, as its weighting mechanism amplifies the noise causing data distortion. On the other hand, \textit{Coop-D}, leveraging a conditional diffusion model for denoising, performs better than \textit{Coop} in all channel conditions, which demonstrates the effectiveness of denoising. Moreover, \textit{Coop-D} outperforms \textit{Coop-W} when the channel condition improves but struggles to address severe signal distortions as effectively as \textit{Coop-W} when the SNR decreases to 0 dB. By integrating adaptive weighting with diffusion-based denoising, \textit{Coop-WD} combines the strengths of both approaches. It mitigates severe channel distortions through weighting while leveraging the diffusion model to reconstruct features and compensate for light distortion. This synergy enables \textit{Coop-WD} to consistently outperform both \textit{Coop-W} and \textit{Coop-D} across all SNR levels. Similar trends and results can also be observed in the realistic WINNER II channel and the more complex non-stationary V2V channel.

To evaluate the scalability of the proposed joint weighting and denoising method with other fusion techniques, Table \ref{table2} presents the average precision using the attentive fusion approach from \cite{openv2v}. At an SNR of 0 dB, the baseline CNN-based autoencoder (AE) shows a significant performance degradation. In contrast, models incorporating denoising and weighting are able to mitigate distortion effects, maintaining accuracy above 45\% at IoU = 0.5 and 26\% at IoU = 0.7. Under moderate distortion conditions (SNR between 10 and 20 dB), the joint weighting and denoising model consistently achieves the highest performance, followed by the model with denoising only. At an SNR of 30 dB, denoising continues to provide performance gains, while the contribution of weighting becomes negligible.

\hlbreakable{To further evaluate the generalization capability of the proposed framework, we integrate \textit{Coop-WD} with two cooperative perception fusion methods, V2X-ViT~\cite{v2x-vit} and Where2comm~\cite{where2comm}, on the V2XSet dataset~\cite{v2x-vit}. The results are reported in Table~\ref{tableXX}. Under the non-stationary V2V channel, \textit{Coop-WD} consistently improves the detection performance of both fusion methods across different SNR levels. For example, at SNR = 0 dB, the AP@0.7 of V2X-ViT increases from 0.205 to 0.497, while that of Where2comm increases from 0.008 to 0.476 after integrating \textit{Coop-WD}. Similar improvements are also observed at SNR = 10 dB and 20 dB. These results demonstrate that the proposed weighting and diffusion-based denoising modules can be used as a plug-in channel-aware enhancement module for different cooperative perception frameworks.}

\subsection{Performance under imperfect CSI, path loss, and time-varying distortions}
\begin{table}[!ht]
\caption{Performance under imperfect CSI.}
\centering
\label{table3}\renewcommand{\arraystretch}{0.9}
\setlength{\tabcolsep}{4pt}
\begin{tabular}{c|cccccc}
\toprule
   \multicolumn{7}{c}{Non-stationary V2V channel with imperfect CSI} 

\\ \midrule
 SNR &
  \multicolumn{2}{c|}{0 dB} &
  \multicolumn{2}{c|}{10 dB} &
  \multicolumn{2}{c}{20 dB} 

   \\ \midrule
 IoU &
  \multicolumn{1}{c}{@0.5} &
  \multicolumn{1}{c|}{@0.7} &
  \multicolumn{1}{c}{@0.5} &
  \multicolumn{1}{c|}{@0.7} &
  \multicolumn{1}{c}{@0.5} &
  \multicolumn{1}{c}{@0.7} 

   \\ 
    \midrule
    Coop &
    \multicolumn{1}{c}{0.026} &
  \multicolumn{1}{c|}{0.009} &
  \multicolumn{1}{c}{0.102} &
  \multicolumn{1}{c|}{0.044} &
  \multicolumn{1}{c}{0.318} &
  \multicolumn{1}{c}{0.153} 
  \\ \midrule
    Coop-W &
    \multicolumn{1}{c}{\underline{0.390}} &
  \multicolumn{1}{c|}{\underline{0.156}} &
  \multicolumn{1}{c}{\underline{0.351}} &
  \multicolumn{1}{c|}{0.139} &
  \multicolumn{1}{c}{0.397} &
  \multicolumn{1}{c}{0.116} 

  \\ \midrule
    Coop-D &
  \multicolumn{1}{c}{0.118} &
  \multicolumn{1}{c|}{0.051} &
  \multicolumn{1}{c}{0.323} &
  \multicolumn{1}{c|}{\underline{0.152}} &
  \multicolumn{1}{c}{\underline{0.519}} &
  \multicolumn{1}{c}{\underline{0.258}}

    \\ \midrule
    Coop-WD &
  \multicolumn{1}{c}{\textbf{0.450}} &
  \multicolumn{1}{c|}{\textbf{0.226}} &
  \multicolumn{1}{c}{\textbf{0.551}} &
  \multicolumn{1}{c|}{\textbf{0.258}} &
  \multicolumn{1}{c}{\textbf{0.600}} &
  \multicolumn{1}{c}{\textbf{0.297}} 
  
   \\
   \bottomrule
\end{tabular}
\end{table}

\begin{figure*}[htb] 
    \centering
    \begin{subfigure}{0.32\textwidth} %, height=3cm
    \centering
        \includegraphics[width=\textwidth]{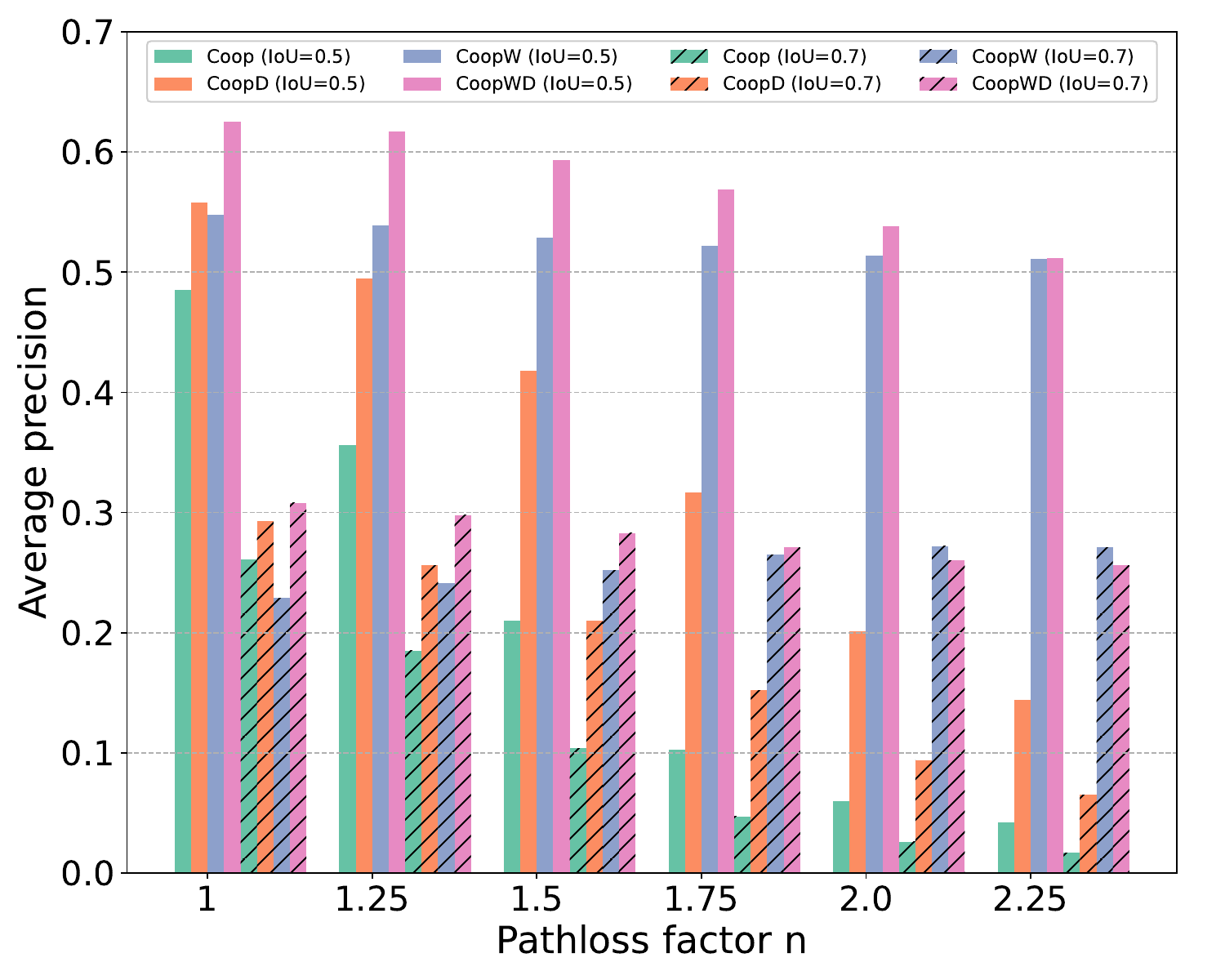}
        \caption{}
        \label{fig:pathloss2}
    \end{subfigure}
    \hfill
 \begin{subfigure}{0.32\textwidth} %, height=3cm
    \centering
        \includegraphics[width=\textwidth]{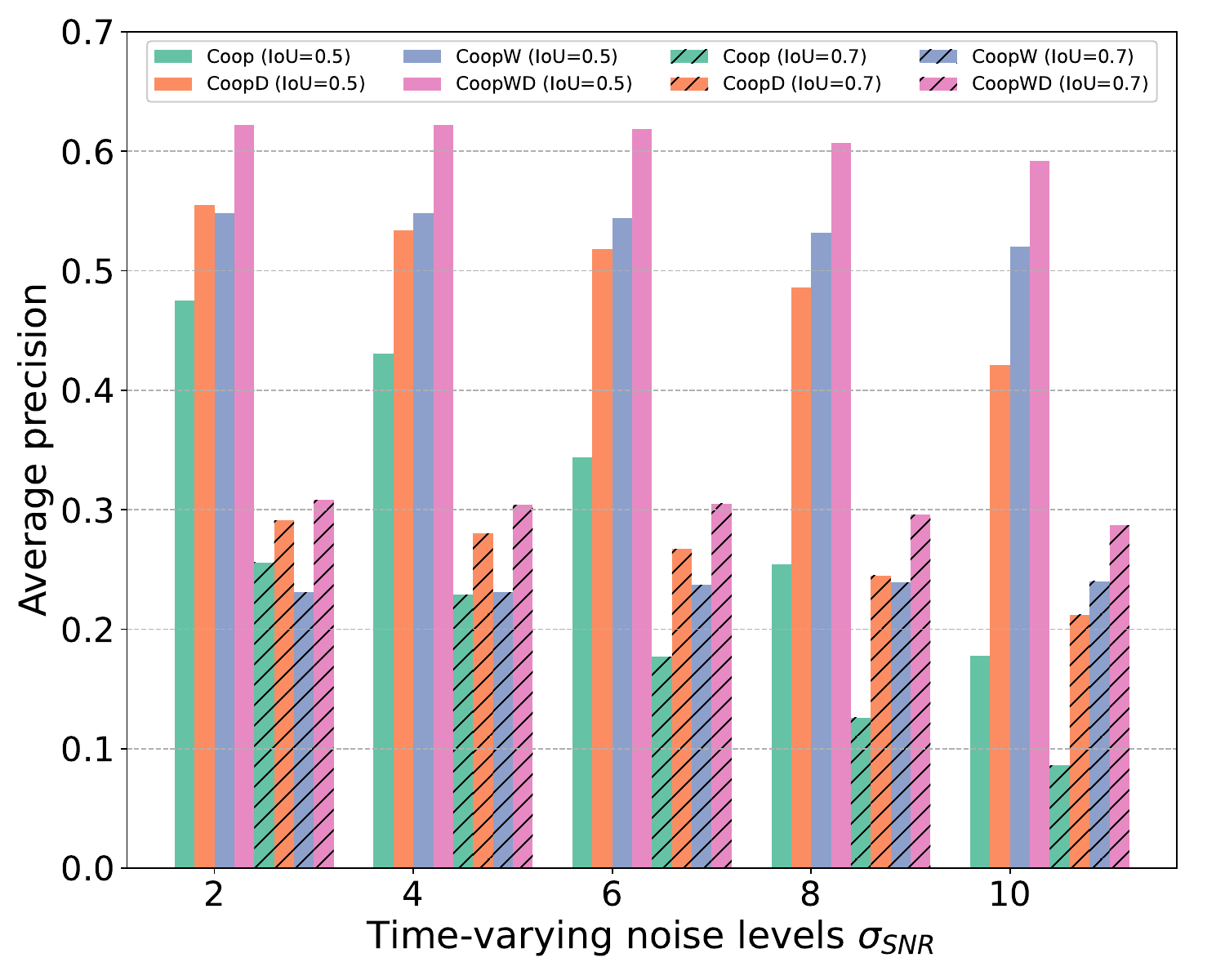}
        \caption{}
        \label{fig:snr_std}
    \end{subfigure}
    \hfill
    \begin{subfigure}{0.32\textwidth} %, height=3cm
    \centering
        \includegraphics[width=\textwidth]{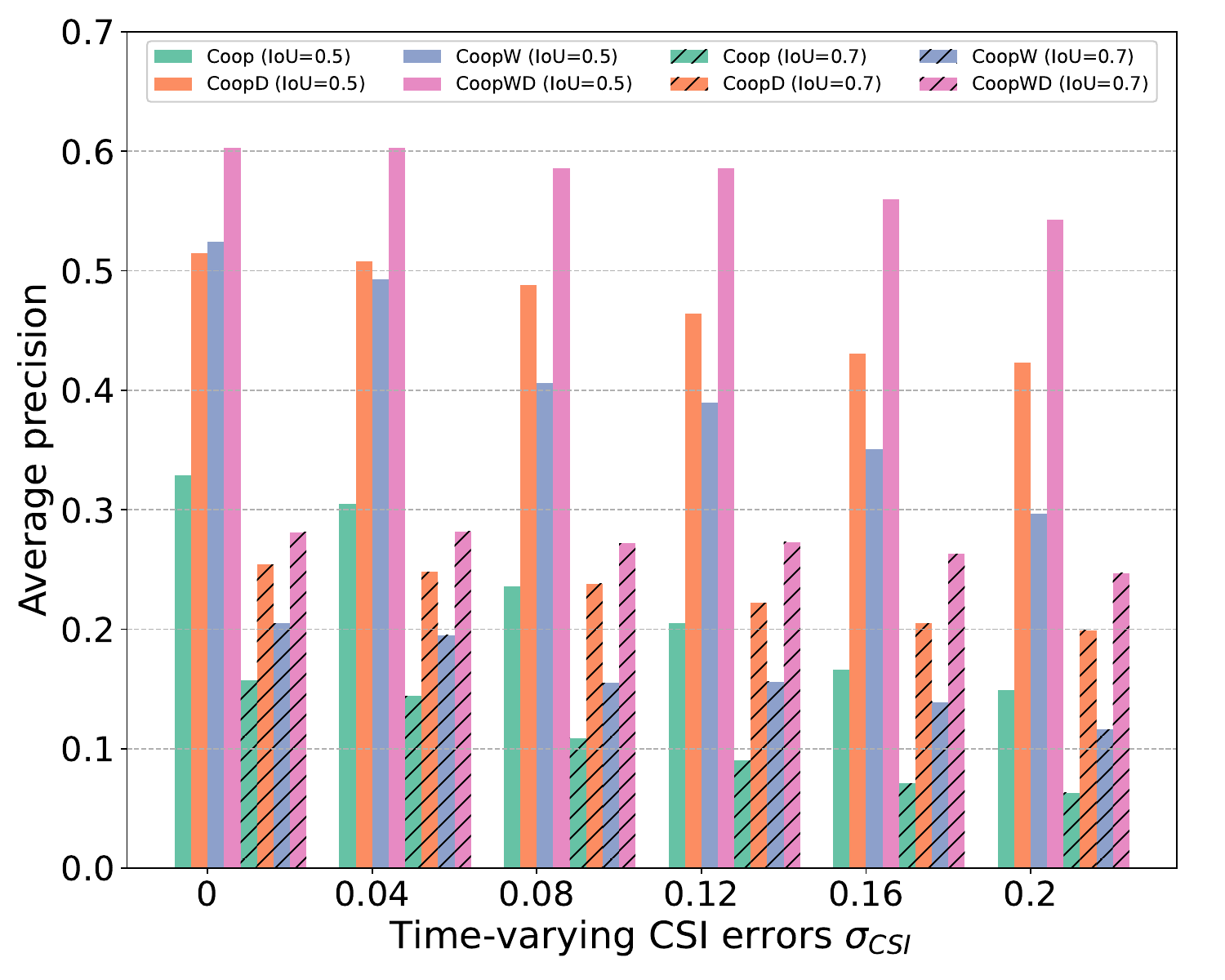}
        \caption{}
        \label{fig:CSI_std}
    \end{subfigure}
    \hfill
    \caption{Performance under different path loss factors and time-varying disturbances. Disturbances are simulated with a fixed time duration following Gaussian distribution. (a) Path loss factor $n$. (b) Time-varying noise levels ($\sigma_\text{SNR}$). (c) Time-varying CSI errors($\sigma_\text{CSI}$).}
    \label{fig:std}
\end{figure*}
In this section, the effects of different path loss factors and time-varying disturbances are evaluated.

Table~\ref{table3} shows the performance of cooperative perception under imperfect CSI. It is demonstrated that the proposed \textit{Coop-WD} performs better than the baseline models \textit{Coop} and \textit{Coop-W} in the presence of CSI error when the SNR ranges from 0 to 20 dB.   

Fig.~\ref{fig:std}(\subref{fig:pathloss2}) shows the performance of the cooperative perception for different path loss factors when the SNR is 30 dB in the non-stationary V2V channel. 
When the path loss factor increases from 1.0 to 2.25, the baseline and Coop-D approach experience a substantial decline in accuracy, decreasing from approximately 55\%, 25\% to less than 10\% for IoU=0.5 and 0.7, respectively. However, \textit{Coop-W} and the \textit{Coop-WD} demonstrate better robustness to changing path loss factors, where \textit{Coop-WD} has a higher accuracy than \textit{Coop-W} due to additional diffusion-based denoising module. It is validated that the proposed joint approach can enhance the system robustness to various communications environments. 

Furthermore, we evaluate the performance of the proposed method under time-varying distortions. Specifically, we assume that the shared feature map is divided by multiple time slots for transmission. During each time slot, Gaussian disturbances are simulated separately for CSI errors and noise levels to model time-varying dynamic distortions. Fig. \ref{fig:std}(\subref{fig:snr_std}) shows the cooperative perception performance under the effects of noise variations over time. As $\sigma_\text{SNR}$ increases to 10, the baseline, \textit{Coop}, experiences a substantial performance drop, retaining less than 50\% of the original AP scores due to the noise variation. In contrast, the proposed \textit{Coop-WD} demonstrates a performance decline of less than 5\% in AP, showing its significantly better robustness against time-varying noise variations. Similar trends and conclusions can be obtained for the variations of CSI errors in Fig. \ref{fig:std}(\subref{fig:CSI_std}). 

\subsection{Model complexity}
\begin{table}[!ht]
\caption{Comparison of average runtime and average precision of Coop-WD with fast-sample schedule, full step for inference and ResShift \cite{10681246}.}
\centering
\label{tableXX}\renewcommand{\arraystretch}{0.9}
\setlength{\tabcolsep}{1pt}
\begin{tabular}{c|cccccc}
\toprule
   \multicolumn{7}{c}{Non-stationary V2V channel} 
\\
\toprule
  SNR&
  \multicolumn{3}{c|}{0 dB} &
  \multicolumn{3}{c}{20 dB} 

   \\ \midrule
 IoU&
  \multicolumn{1}{c}{AP@0.5} &
  \multicolumn{1}{c}{AP@0.7} &
  \multicolumn{1}{c|}{Runtime} &
  \multicolumn{1}{c}{AP@0.5} &
  \multicolumn{1}{c}{AP@0.7} &
  \multicolumn{1}{c}{Runtime} 

  \\ \midrule
    Fast-sample schedule&
  \multicolumn{1}{c}{0.487} &
  \multicolumn{1}{c}{0.260} &
  \multicolumn{1}{c|}{68.0ms} &
  \multicolumn{1}{c}{0.617} &
  \multicolumn{1}{c}{0.309} &
  \multicolumn{1}{c}{68.2ms} \\
  \midrule

  Full-step inference &
  \multicolumn{1}{c}{0.479} &
  \multicolumn{1}{c}{0.278} &
  \multicolumn{1}{c|}{332.5ms} &
  \multicolumn{1}{c}{0.609} &
  \multicolumn{1}{c}{0.312} &
  \multicolumn{1}{c}{324.9ms} \\

  \midrule

  ResShift \cite{10681246} &
  \multicolumn{1}{c}{0.465} &
  \multicolumn{1}{c}{0.280} &
  \multicolumn{1}{c|}{125.3ms} &
  \multicolumn{1}{c}{0.475} &
  \multicolumn{1}{c}{0.305} &
  \multicolumn{1}{c}{125.6ms} 

    \\ \bottomrule
\end{tabular}
\end{table}

\begin{table}[!ht]
\caption{Comparison of average runtime, model size and memory usage of \textit{Coop}, \textit{Coop-W}, \textit{Coop-D}, and \textit{Coop-WD}.}
\label{complexity} \renewcommand{\arraystretch}{0.9}
\centering
\begin{tabular}{cccc}
\toprule 

        \multirow{2}{*}{}& \multirow{2}{*}{\shortstack{Average Runtime\\ (ms)}} & \multirow{2}{*}{\shortstack{Model Size\\ (MB)}} & \multirow{2}{*}{\shortstack{Memory \\ (MB)}}\\ 
        & & & \\
        \midrule
Coop    & 24.3               & 63.4    &2921.8         \\ 
Coop-W  & 26.2                & 63.4   &2929.9         \\ 
Coop-D & 66.2               & 135.9    &8507.1      \\ 
Coop-WD & 69.0               & 135.9   &8523.9        \\
\bottomrule
\end{tabular}
\end{table}

\begin{table}[!ht]
\caption{Comparison of average runtime and average precision of Coop-WD and Coop-WD-eco for each communication channel.}
\centering
\label{table5}\renewcommand{\arraystretch}{0.9}
\setlength{\tabcolsep}{2.5pt}
\begin{tabular}{c|cccccc}
\toprule
   \multicolumn{7}{c}{Non-stationary V2V channel} 

\\ \midrule
 SNR &
  \multicolumn{3}{c|}{0 dB} &
  \multicolumn{3}{c}{20 dB} 

   \\ \midrule
 IoU &
  \multicolumn{1}{c}{AP@0.5} &
  \multicolumn{1}{c}{AP@0.7} &
  \multicolumn{1}{c|}{Runtime} &
  \multicolumn{1}{c}{AP@0.5} &
  \multicolumn{1}{c}{AP@0.7} &
  \multicolumn{1}{c}{Runtime} 

  \\ \midrule
    Coop-WD &
  \multicolumn{1}{c}{0.487} &
  \multicolumn{1}{c}{0.260} &
  \multicolumn{1}{c|}{69.1ms} &
  \multicolumn{1}{c}{0.617} &
  \multicolumn{1}{c}{0.309} &
  \multicolumn{1}{c}{69.2ms} \\
  \midrule

  Coop-WD-eco &
  \multicolumn{1}{c}{0.481} &
  \multicolumn{1}{c}{0.253} &
  \multicolumn{1}{c|}{26.9ms} &
  \multicolumn{1}{c}{0.616} &
  \multicolumn{1}{c}{0.306} &
  \multicolumn{1}{c}{59.3ms}

    \\ \midrule

  \multicolumn{7}{c}{Rician fading channel}   
  \\ \midrule
    Coop-WD &
  \multicolumn{1}{c}{0.483} &
  \multicolumn{1}{c}{0.264} &
  \multicolumn{1}{c|}{68.7ms} &
  \multicolumn{1}{c}{0.549} &
  \multicolumn{1}{c}{0.286} &
  \multicolumn{1}{c}{68.5ms} \\
  \midrule

 Coop-WD-eco &
  \multicolumn{1}{c}{0.491} &
  \multicolumn{1}{c}{0.273} &
  \multicolumn{1}{c|}{33.7ms} &
  \multicolumn{1}{c}{0.548} &
  \multicolumn{1}{c}{0.283} &
  \multicolumn{1}{c}{60.8ms}

    \\ \midrule
  
  \multicolumn{7}{c}{WINNER II channel}   
  \\ \midrule
    Coop-WD &
  \multicolumn{1}{c}{0.462} &
  \multicolumn{1}{c}{0.266} &
  \multicolumn{1}{c|}{69ms} &
  \multicolumn{1}{c}{0.623} &
  \multicolumn{1}{c}{0.330} &
  \multicolumn{1}{c}{69ms}

   \\
   \midrule
    Coop-WD-eco &
  \multicolumn{1}{c}{0.433} &
  \multicolumn{1}{c}{0.229} &
  \multicolumn{1}{c|}{34.1ms} &
  \multicolumn{1}{c}{0.620} &
  \multicolumn{1}{c}{0.312} &
  \multicolumn{1}{c}{68.9ms}

    \\ \bottomrule
\end{tabular}
\end{table}

\begin{figure}[!ht]
    \centering
   
    \subfloat[]{
        \includegraphics[width=0.23\textwidth]{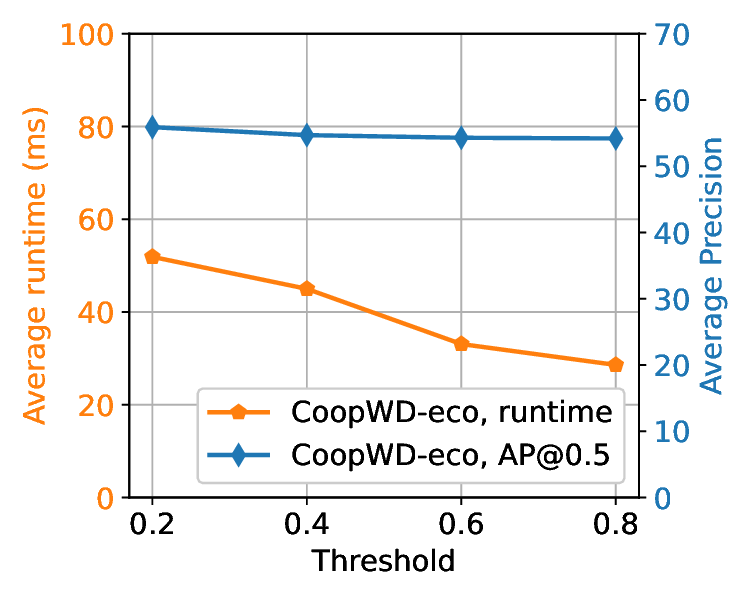}
        \label{Fig.runtime_threshold}}
    \hfill
    \subfloat[]{
        \includegraphics[width=0.23\textwidth]{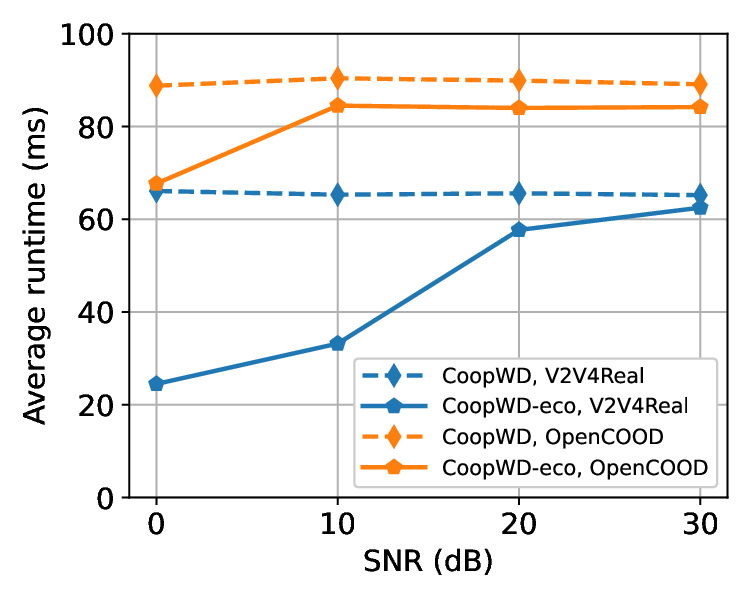}
        \label{Fig.runtime_snr}}
    \caption{Runtime analysis of Coop-WD and Coop-WD-eco in the non-stationary V2V channel. \hlbreakable{(a) Average runtime of Coop-WD-eco under different weighting-output thresholds at 10 dB SNR on the V2V4Real dataset.} (b) Average runtime across SNRs from 0 to 30 dB on the V2V4Real and OPV2V datasets.}
    \label{Fig.runtime}
\end{figure}

In Table \ref{tableXX}, we first evaluate the impact of diffusion step settings in Coop-WD under the non-stationary V2V channel by comparing full 50-step inference with the fast-sampling schedule using 5 steps. The difference in average precision remains within 1-2\% across IoU thresholds, with only slight improvement at IoU=0.7 under 50-step inference due to additional iterative denoising steps. \hlbreakable{However, 50-step inference increases the runtime from approximately 68 ms to over 330 ms, corresponding to nearly a 5-time increase in latency, which is unsuitable for latency-sensitive cooperative perception. In contrast, 5-step inference reduce the runtime below 100 ms while preserving comparable detection accuracy. Therefore, the 5-step fast-sampling schedule provides a practical accuracy-efficiency trade-off and is adopted in all \textit{Coop-WD} experiments.}

To further demonstrate the compatibility of the proposed framework with existing diffusion-based reconstruction models, we also implement ResShift \cite{10681246} under the same settings. ResShift adopts a residual-based schedule designed to accelerate image restoration while preserving reconstruction quality. As shown in Table \ref{tableXX}, at SNR = 0 dB, ResShift achieves slightly higher AP at IoU=0.7 compared with the fast-sampling Coop-WD, indicating that residual-based reconstruction can effectively recover fine-grained structures under severe distortion. However, at SNR = 20 dB, Coop-WD achieves higher AP at IoU=0.5 and comparable AP at IoU=0.7 with significantly lower runtime.

Table \ref{complexity} compares each approach's average runtime, model size, and memory usage. Compared with the baseline without weighting and denoising modules, \textit{Coop-W} increases runtime slightly to 26.2 ms with minimal impact on model size; \textit{Coop-D} raises runtime to 66.2 ms and doubles the model size to 135.9 MB due to additional diffusion-based pixel-level feature processing. A similar trend is observed in CUDA memory usage, where the inclusion of the diffusion model results in roughly a threefold increase from about 2900 MB to over 8500 MB in memory consumption. This additional overhead is due to multi-step diffusion-based reconstruction on full-resolution feature maps, which enables recovery of channel-induced distortions that cannot be handled by CAV-level weighting alone.

Although reducing model complexity is not the primary focus of this work, \textit{Coop-WD} achieves the best detection performance among the benchmarks with a higher computational cost introduced by the additional diffusion model. To reduce this cost, we use the output of the weighting model as an indicator of feature reliability. This output usually approaches 0 under severe distortion and approaches 1 as the channel condition improves. As shown in Table~\ref{table1}, the denoising module is more effective under light distortion, but brings limited improvement under severe distortion compared with weighting alone. Based on this observation, we propose an efficient variant, \textit{Coop-WD-eco}, which activates the diffusion-based denoising module only when the received features are relatively reliable. Specifically, when the weighting output falls below the threshold of 0.6, the system bypasses the denoising process.

The performance comparison between \textit{Coop-WD} and \textit{Coop-WD-eco} is shown in Table~\ref{table5}. By using the threshold of 0.6 to disable denoising under severe distortion, \textit{Coop-WD-eco} reduces computational cost by more than 50\% compared with \textit{Coop-WD}. At higher SNR levels, such as 20 dB, \textit{Coop-WD-eco} trades a slight decrease in precision for about a 10\% improvement in computational efficiency. \hlbreakable{To validate the threshold selection, Fig.\ref{Fig.runtime_threshold} shows the runtime and average precision of \textit{Coop-WD-eco} under different thresholds at 10 dB SNR on the V2V4Real dataset. A larger threshold bypasses denoising more often and reduces runtime, while the average precision changes only slightly. This result suggests that \textit{Coop-WD-eco} is not highly sensitive to the threshold, and that 0.6 gives a stable accuracy--efficiency trade-off.}

We also evaluated the inference time under the non-stationary V2V channel across SNRs from 0 to 30 dB, as shown in Fig. \ref{Fig.runtime_snr}. The proposed models were tested on both the V2V4Real and OPV2V datasets. Overall, \textit{Coop-WD-eco} runs faster than \textit{Coop-WD}, although the gap becomes smaller as the SNR increases. The runtime of \textit{Coop-WD-eco} increases with channel quality, ranging from about 24 ms to 60 ms on V2V4Real and from 69 ms to 84 ms on OPV2V. This trend comes from the adaptive use of the denoising module. Under severe distortion at low SNR, denoising is deactivated according to the weighting score because the expected performance gain is limited. Under light distortion at high SNR, denoising is activated to refine fine-grained features, which increases inference time. In contrast, the average runtime of \textit{Coop-WD} remains relatively stable, at approximately 67 ms for V2V4Real and 90 ms for OPV2V, regardless of the SNR.

\begin{figure}[!ht]
    \centering
    \subfloat[]{
        \includegraphics[width=0.48\textwidth]{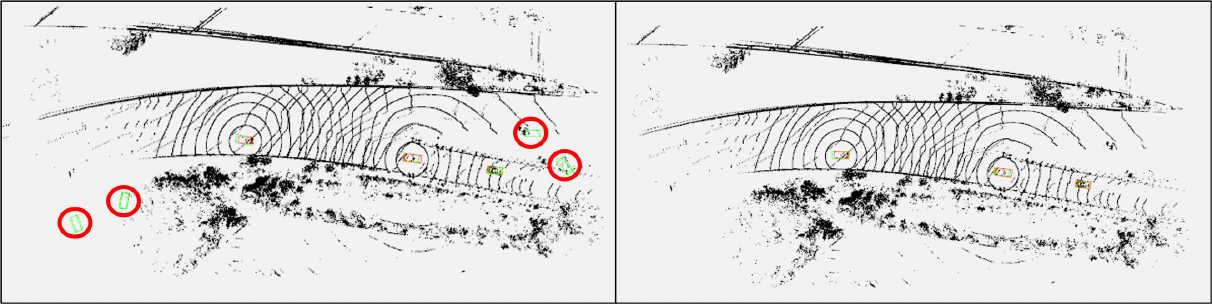}
        \label{Fig.FP1}}
    \hfill
    \subfloat[]{
    \includegraphics[width=0.48\textwidth]{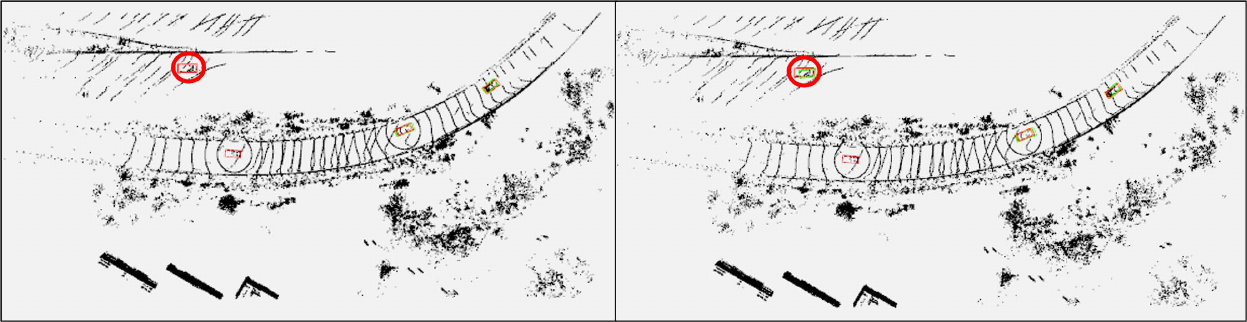}
        \label{Fig.FN1}}
    \caption{Visualization examples of Coop-D (left) and the Coop-WD (right). (a) False positive correction. (b) False negative correction.}
    \label{Fig.d_dw}
\end{figure}

\begin{figure}[!ht]
    \centering
    \subfloat[]{\includegraphics[width=0.48\textwidth]{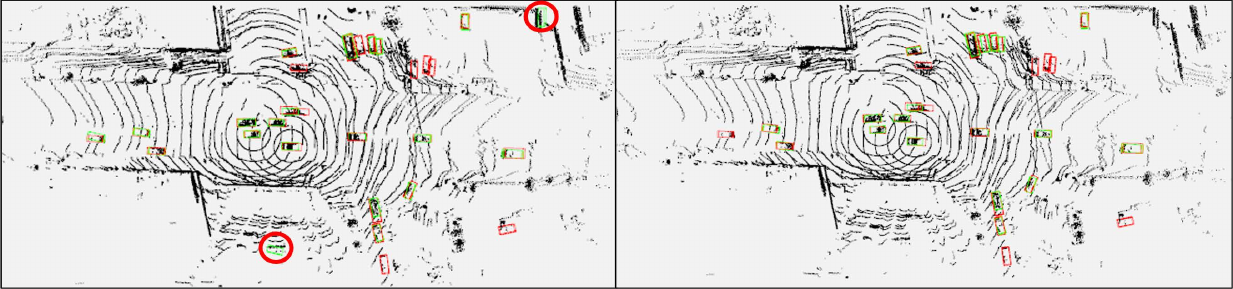}
        \label{Fig.FP2}}

    \hfill
    \subfloat[]{\includegraphics[width=0.48\textwidth]{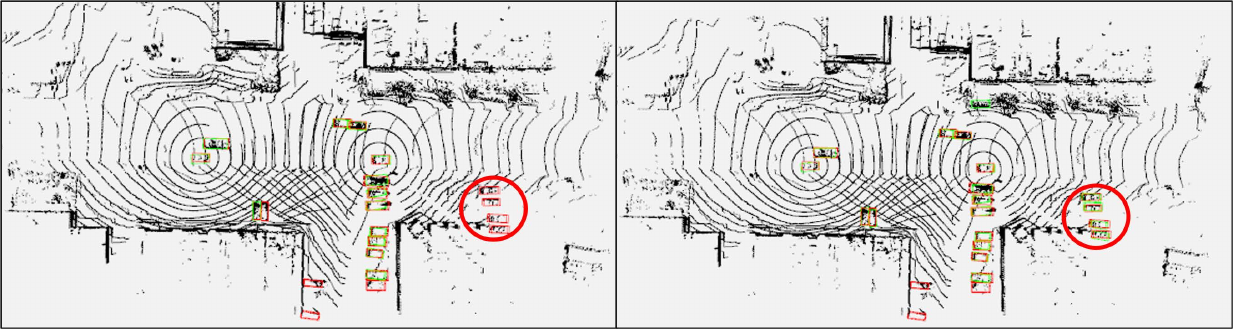}
        \label{Fig.FN2}}
    \caption{Visualization examples of Coop-W (left) and Coop-WD (right). (a) False positive correction. (b) False negative correction.}
    \label{Fig.w_dw}
\end{figure}

\begin{figure}[!ht]
    \centering
    \subfloat[]{\includegraphics[width=0.48\textwidth]{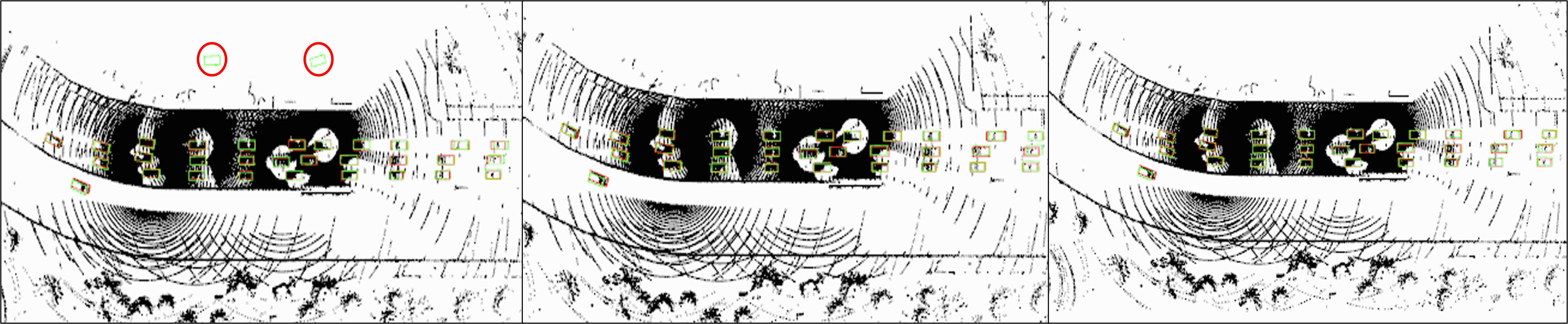}
        \label{Fig.FP3}}

    \hfill
    \subfloat[]{\includegraphics[width=0.48\textwidth]{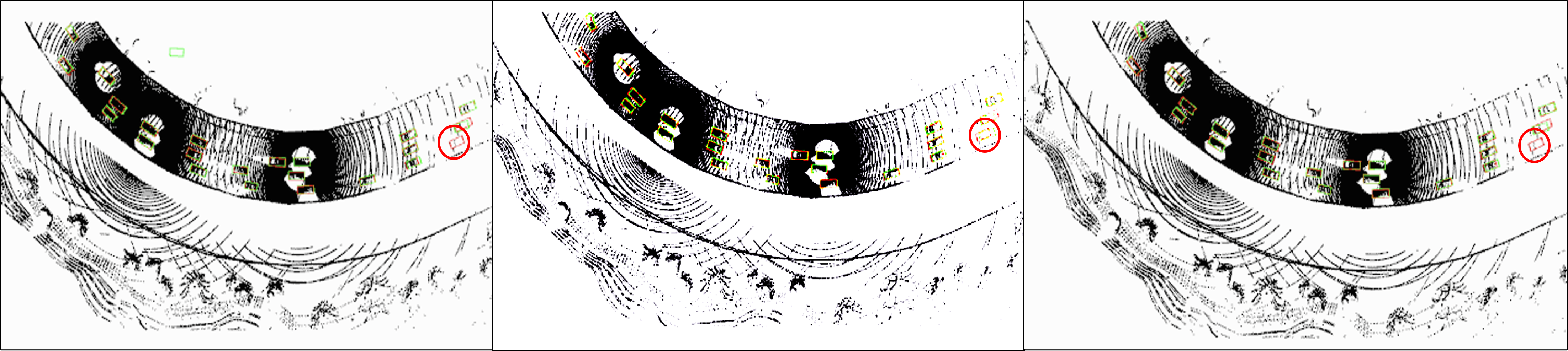}
        \label{Fig.FN3}}
    \caption{Visualization examples of \textit{Coop} (left), \textit{Coop-WD} (middle), and \textit{Coop-WD-eco} (right). (a) False positive correction. (b) False negative correction.}
    \label{Fig.opencood}
\end{figure}

\subsection{Qualitative analysis}
In this section, qualitative analysis with visual examples is provided for the ablation study to validate the effectiveness of the proposed method in Figs. \ref{Fig.d_dw}, \ref{Fig.w_dw} and \ref{Fig.opencood}, where the bounding boxes of ground truth and detection results are marked as red and green, respectively. 
\begin{itemize}
    \item \textbf{Comparison of the pixel-level \textit{Coop-D} and the joint \textit{Coop-WD}:} Fig. \ref{Fig.d_dw} compares the performance of \textit{Coop-D} and \textit{Coop-WD} on the same data frame. In Fig. \ref{Fig.d_dw} (a), \textit{Coop-D} has false positive predictions, denoted by the red circles, due to channel distortions. However, the proposed \textit{Coop-WD} can effectively avoid these false positive predictions. In Fig. \ref{Fig.d_dw}(b), \textit{Coop-D} fails to detect a moving object denoted by the red circle, while the proposed \textit{Coop-WD} successfully identifies it.
    
    \item \textbf{Comparison of the CAV-level \textit{Coop-W} and the joint \textit{Coop-WD}:} Fig. \ref{Fig.w_dw} illustrates the visualization comparison of the same data frame between \textit{Coop-W} and the proposed \textit{Coop-WD}. \textit{Coop-D} falsely identifies some areas without actual objects as targets in Fig. \ref{Fig.w_dw} (a) and failed to detect some targets in Fig. \ref{Fig.w_dw} (b). However, \textit{Coop-WD} has effectively addressed these false detections or undetected results.

    \item \textbf{Comparison of the \textit{Coop}, \textit{Coop-WD}, and \textit{Coop-WD-eco} for multiple CAVs:} Fig. \ref{Fig.opencood} shows the visualization examples with multiple CAVs using the OPV2V dataset \cite{openv2v}. The results demonstrate that both \textit{Coop-WD} and \textit{Coop-WD-eco} could effectively reduce false-positive detections and correct false detections where the baseline \textit{Coop} fails. This further validates the scalability and effectiveness of the proposed methods across different data domains in multi-vehicle scenarios.
\end{itemize}

\subsection{Performance under different numbers of CAVs}

\begin{table}[!ht]
\caption{Comparison of average runtime of \textit{Coop-WD} and \textit{Coop-WD-eco} for different number of CAVs on the OPV2V dataset \cite{openv2v}.}
\centering
\label{table6}\renewcommand{\arraystretch}{0.9}
\setlength{\tabcolsep}{2.5pt}
\begin{tabular}{c|ccccccccc}
\toprule
   \multicolumn{9}{c}{Average Runtime (ms)} 

\\ \midrule
 SNR &
  \multicolumn{4}{c|}{0 dB} &
  \multicolumn{4}{c}{20 dB} 

   \\ \midrule
 CAV number &
  \multicolumn{1}{c}{2} &
  \multicolumn{1}{c}{3} &
  \multicolumn{1}{c}{4} &
  \multicolumn{1}{c|}{5} &
  \multicolumn{1}{c}{2} &
  \multicolumn{1}{c}{3} &
  \multicolumn{1}{c}{4} &
  \multicolumn{1}{c}{5} 

  \\ \midrule
    Coop-WD &
  \multicolumn{1}{c}{71.8} &
  \multicolumn{1}{c}{103.1} &
  \multicolumn{1}{c}{136.5} &
  \multicolumn{1}{c|}{165.8} &
  \multicolumn{1}{c}{69.4} &
  \multicolumn{1}{c}{103.0}&
  \multicolumn{1}{c}{138.5}&
  \multicolumn{1}{c}{169.4} \\
  \midrule

  Coop-WD-eco &
  \multicolumn{1}{c}{54.4} &
  \multicolumn{1}{c}{77.0} &
  \multicolumn{1}{c}{104.5} &
  \multicolumn{1}{c|}{128.9} &
  \multicolumn{1}{c}{65.9} &
  \multicolumn{1}{c}{98.8}&
  \multicolumn{1}{c}{133.6}&
  \multicolumn{1}{c}{163.3}
  \\

 \bottomrule
\end{tabular}
\end{table}

\begin{table}[!ht]
\caption{Comparison of average precision of \textit{Coop-WD} and \textit{Coop-WD-eco} for different number of CAVs on the OPV2V dataset \cite{openv2v}.}
\centering
\label{table7}\renewcommand{\arraystretch}{0.9}
\setlength{\tabcolsep}{2.5pt}
\begin{tabular}{c|ccccccccc}
\toprule
   \multicolumn{9}{c}{Average Precision} 

   \\ \midrule
 CAV number &
  \multicolumn{2}{c|}{2} &
  \multicolumn{2}{c|}{3} &
  \multicolumn{2}{c|}{4} &
  \multicolumn{2}{c}{5} 
  \\
  \midrule
 IoU &
  \multicolumn{1}{c}{@0.5} &
  \multicolumn{1}{c|}{@0.7} &
  \multicolumn{1}{c}{@0.5} &
  \multicolumn{1}{c|}{@0.7} &
  \multicolumn{1}{c}{@0.5} &
  \multicolumn{1}{c|}{@0.7} &
  \multicolumn{1}{c}{@0.5} &
  \multicolumn{1}{c}{@0.7} 
  
  \\ \midrule
 SNR &
  \multicolumn{8}{c}{0 dB} 
  \\ \midrule
    Coop-WD &
  \multicolumn{1}{c}{59.8} &
  \multicolumn{1}{c|}{43.8} &
  \multicolumn{1}{c}{67.4} &
  \multicolumn{1}{c|}{47.9} &
  \multicolumn{1}{c}{68.8} &
  \multicolumn{1}{c|}{45.7}&
  \multicolumn{1}{c}{68.3}&
  \multicolumn{1}{c}{49.5} \\
  \midrule

  Coop-WD-eco &
  \multicolumn{1}{c}{61.0} &
  \multicolumn{1}{c|}{46.4} &
  \multicolumn{1}{c}{64.1} &
  \multicolumn{1}{c|}{46.6} &
  \multicolumn{1}{c}{72.3} &
  \multicolumn{1}{c|}{52.6}&
  \multicolumn{1}{c}{73.1}&
  \multicolumn{1}{c}{55.7}
  \\

  \midrule
   SNR &
  \multicolumn{8}{c}{20 dB} 
  \\ \midrule
    Coop-WD &
  \multicolumn{1}{c}{80.0} &
  \multicolumn{1}{c|}{64.5} &
  \multicolumn{1}{c}{88.0} &
  \multicolumn{1}{c|}{69.7} &
  \multicolumn{1}{c}{84.2} &
  \multicolumn{1}{c|}{66.1}&
  \multicolumn{1}{c}{88.1}&
  \multicolumn{1}{c}{71.3} \\
  \midrule

  Coop-WD-eco &
  \multicolumn{1}{c}{80.1} &
  \multicolumn{1}{c|}{64.4} &
  \multicolumn{1}{c}{88.1} &
  \multicolumn{1}{c|}{70.3} &
  \multicolumn{1}{c}{84.7} &
  \multicolumn{1}{c|}{65.1}&
  \multicolumn{1}{c}{88.9}&
  \multicolumn{1}{c}{71.4}
  \\

 \bottomrule
\end{tabular}
\end{table}

Tables \ref{table6} and \ref{table7} present the average precision and inference runtime under varying numbers of collaborating CAVs. The experiments are conducted on the OPV2V dataset under the OpenCOOD framework \cite{openv2v}. Performance is analyzed by grouping test scenarios according to the number of participating CAVs (i.e., 2–5 vehicles), which aims to illustrate the performance trend as the number of collaborators increases.

In Table \ref{table6}, \textit{Coop-WD-eco} demonstrates comparable computational efficiency to Coop-WD. As the number of CAVs increases from 2 to 5, the inference time for \textit{Coop-WD} rises from approximately 70 ms to 165 ms at both SNR = 0 dB and 20 dB, indicating an average increase of roughly 30 ms per additional CAV. In contrast, Coop-WD-eco achieves around a 30\% reduction in runtime at SNR = 0 dB, and about a 5\% reduction at SNR = 20 dB, showing its improved efficiency, especially in low-SNR conditions. In Table \ref{table7}, although \textit{Coop-WD-eco} deactivates the denoising module when the weighting score falls below the threshold of 0.6, it still achieves performance comparable to \textit{Coop-WD}. At SNR = 0 dB, \textit{Coop-WD-eco} outperforms \textit{Coop-WD} when the number of CAVs exceeds 3, demonstrating that the denoising module may offer limited improvements under severe distortion, particularly in multi-vehicle scenarios. This result indicates that adaptive deactivation of the denoising module can be both computationally efficient and effective in challenging communication environments.

\section{Conclusion and future works}
In this work, we proposed \textit{Coop-WD}, a joint weighting and denoising framework for improving cooperative perception under V2V channel impairments. The framework combines self-supervised feature weighting with generative diffusion-based denoising in a hierarchical manner. Specifically, it enhances transmitted features at both the CAV level and the pixel level, reducing the impact of channel-induced distortions at different granularities. To reduce the additional computational cost of the denoising module, we further introduced \textit{Coop-WD-eco}, an efficient variant that selectively deactivates denoising under severe distortion. The proposed framework was evaluated under several channel models, including Rician fading, WINNER II, and non-stationary V2V channels. Simulation results show that \textit{Coop-WD} consistently outperforms its standalone components, \textit{Coop-W} and \textit{Coop-D}, as well as conventional baselines without weighting or denoising. The results also show that \textit{Coop-WD} mitigates severe signal distortions and improves performance under mild impairments by restoring fine-grained feature information. In addition, \textit{Coop-WD-eco} achieves up to 50\% computational savings under severe channel conditions while maintaining competitive precision as channel quality improves. Qualitative visualizations further confirm the robustness and effectiveness of \textit{Coop-WD} compared with \textit{Coop-W} and \textit{Coop-D}. 

Although the proposed \textit{Coop-WD} achieves promising numerical results, several limitations remain. \hlbreakable{First, the diffusion-based denoising module increases computational cost, while \textit{Coop-WD-eco} mainly mitigates this under severe distortion. Future work will explore lightweight diffusion architectures, adaptive sampling, and resource-aware gating to further improve efficiency.} Second, generative and self-supervised models may behave unpredictably under out-of-distribution conditions, affecting perception reliability. We will integrate responsible AI principles, develop a communication-aware verification framework to quantify uncertainties from distorted or incomplete V2V information, and design domain adaptation techniques to improve generalization across varying channel and network conditions.
\bibliographystyle{IEEEtran}
% \bibliography{reference.bib}

\begin{thebibliography}{10}
\providecommand{\url}[1]{#1}
\csname url@samestyle\endcsname
\providecommand{\newblock}{\relax}
\providecommand{\bibinfo}[2]{#2}
\providecommand{\BIBentrySTDinterwordspacing}{\spaceskip=0pt\relax}
\providecommand{\BIBentryALTinterwordstretchfactor}{4}
\providecommand{\BIBentryALTinterwordspacing}{\spaceskip=\fontdimen2\font plus
\BIBentryALTinterwordstretchfactor\fontdimen3\font minus \fontdimen4\font\relax}
\providecommand{\BIBforeignlanguage}[2]{{%
\expandafter\ifx\csname l@#1\endcsname\relax
\typeout{** WARNING: IEEEtran.bst: No hyphenation pattern has been}%
\typeout{** loaded for the language `#1'. Using the pattern for}%
\typeout{** the default language instead.}%
\else
\language=\csname l@#1\endcsname
\fi
#2}}
\providecommand{\BIBdecl}{\relax}
\BIBdecl

\bibitem{pointpillars}
A.~H. Lang, S.~Vora, H.~Caesar, L.~Zhou, J.~Yang, and O.~Beijbom, ``Pointpillars: Fast encoders for object detection from point clouds,'' in \emph{Proceedings of the IEEE/CVF conference on computer vision and pattern recognition}, 2019, pp. 12\,697--12\,705.

\bibitem{shi2020pv}
S.~Shi, C.~Guo, L.~Jiang, Z.~Wang, J.~Shi, X.~Wang, and H.~Li, ``Pv-rcnn: Point-voxel feature set abstraction for 3d object detection,'' in \emph{Proceedings of the IEEE/CVF conference on computer vision and pattern recognition}, 2020, pp. 10\,529--10\,538.

\bibitem{f-cooper}
Q.~Chen, X.~Ma, S.~Tang, J.~Guo, Q.~Yang, and S.~Fu, ``F-cooper: Feature based cooperative perception for autonomous vehicle edge computing system using 3d point clouds,'' in \emph{Proceedings of the 4th ACM/IEEE Symposium on Edge Computing}, 2019, pp. 88--100.

\bibitem{openv2v}
R.~Xu, H.~Xiang, X.~Xia, X.~Han, J.~Li, and J.~Ma, ``Opv2v: An open benchmark dataset and fusion pipeline for perception with vehicle-to-vehicle communication,'' in \emph{2022 International Conference on Robotics and Automation (ICRA)}.\hskip 1em plus 0.5em minus 0.4em\relax IEEE, 2022, pp. 2583--2589.

\bibitem{v2vnet}
T.~Wang, S.~Manivasagam, M.~Liang, B.~Yang, W.~Zeng, and R.~Urtasun, ``V2vnet: Vehicle-to-vehicle communication for joint perception and prediction,'' in \emph{Computer Vision--ECCV 2020: 16th European Conference, Glasgow, UK, August 23--28, 2020, Proceedings, Part II 16}.\hskip 1em plus 0.5em minus 0.4em\relax Springer, 2020, pp. 605--621.

\bibitem{v2x-vit}
R.~Xu, H.~Xiang, Z.~Tu, X.~Xia, M.-H. Yang, and J.~Ma, ``V2x-vit: Vehicle-to-everything cooperative perception with vision transformer,'' in \emph{European conference on computer vision}.\hskip 1em plus 0.5em minus 0.4em\relax Springer, 2022, pp. 107--124.

\bibitem{cobevt}
R.~Xu, Z.~Tu, H.~Xiang, W.~Shao, B.~Zhou, and J.~Ma, ``Cobevt: Cooperative bird’s eye view semantic segmentation with sparse transformers,'' in \emph{Conference on Robot Learning}.\hskip 1em plus 0.5em minus 0.4em\relax PMLR, 2023, pp. 989--1000.

\bibitem{when2com}
Y.-C. Liu, J.~Tian, N.~Glaser, and Z.~Kira, ``When2com: Multi-agent perception via communication graph grouping,'' in \emph{Proceedings of the IEEE/CVF Conference on computer vision and pattern recognition}, 2020, pp. 4106--4115.

\bibitem{where2comm}
Y.~Hu, S.~Fang, Z.~Lei, Y.~Zhong, and S.~Chen, ``Where2comm: Communication-efficient collaborative perception via spatial confidence maps,'' \emph{Advances in neural information processing systems}, vol.~35, pp. 4874--4886, 2022.

\bibitem{who2com}
Y.~Liu, J.~Tian, C.-Y. Ma, N.~Glaser, C.-W. Kuo, and Z.~Kira, ``Who2com: Collaborative perception via learnable handshake communication,'' in \emph{2020 IEEE International Conference on Robotics and Automation (ICRA)}.\hskip 1em plus 0.5em minus 0.4em\relax IEEE, 2020, pp. 6876--6883.

\bibitem{10054381}
C.-X. Wang, X.~You, X.~Gao, X.~Zhu, Z.~Li, C.~Zhang, H.~Wang, Y.~Huang, Y.~Chen, H.~Haas, J.~S. Thompson, E.~G. Larsson, M.~D. Renzo, W.~Tong, P.~Zhu, X.~Shen, H.~V. Poor, and L.~Hanzo, ``On the road to 6g: Visions, requirements, key technologies, and testbeds,'' \emph{IEEE Communications Surveys \& Tutorials}, vol.~25, no.~2, pp. 905--974, 2023.

\bibitem{8869705}
W.~Saad, M.~Bennis, and M.~Chen, ``A vision of 6g wireless systems: Applications, trends, technologies, and open research problems,'' \emph{IEEE Network}, vol.~34, no.~3, pp. 134--142, 2020.

\bibitem{8054694}
T.~O’Shea and J.~Hoydis, ``An introduction to deep learning for the physical layer,'' \emph{IEEE Transactions on Cognitive Communications and Networking}, vol.~3, no.~4, pp. 563--575, 2017.

\bibitem{9733260}
C.~Liu, Y.~Chen, and S.-H. Yang, ``Deep learning based detection for communications systems with radar interference,'' \emph{IEEE Transactions on Vehicular Technology}, vol.~71, no.~6, pp. 6245--6254, 2022.

\bibitem{9398576}
H.~Xie, Z.~Qin, G.~Y. Li, and B.-H. Juang, ``Deep learning enabled semantic communication systems,'' \emph{IEEE Transactions on Signal Processing}, vol.~69, pp. 2663--2675, 2021.

\bibitem{10345474}
C.~Liu, Y.~Zhou, Y.~Chen, and S.-H. Yang, ``Knowledge distillation-based semantic communications for multiple users,'' \emph{IEEE Transactions on Wireless Communications}, vol.~23, no.~7, pp. 7000--7012, 2024.

\bibitem{early1}
E.~Arnold, M.~Dianati, R.~de~Temple, and S.~Fallah, ``Cooperative perception for 3d object detection in driving scenarios using infrastructure sensors,'' \emph{IEEE Transactions on Intelligent Transportation Systems}, vol.~23, no.~3, pp. 1852--1864, 2020.

\bibitem{early2}
H.~Gao, B.~Cheng, J.~Wang, K.~Li, J.~Zhao, and D.~Li, ``Object classification using cnn-based fusion of vision and lidar in autonomous vehicle environment,'' \emph{IEEE Transactions on Industrial Informatics}, vol.~14, no.~9, pp. 4224--4231, 2018.

\bibitem{late1}
M.~Ambrosin, I.~J. Alvarez, C.~Buerkle, L.~L. Yang, F.~Oboril, M.~R. Sastry, and K.~Sivanesan, ``Object-level perception sharing among connected vehicles,'' in \emph{2019 IEEE Intelligent Transportation Systems Conference (ITSC)}.\hskip 1em plus 0.5em minus 0.4em\relax IEEE, 2019, pp. 1566--1573.

\bibitem{late2}
Z.~Zhang, S.~Wang, Y.~Hong, L.~Zhou, and Q.~Hao, ``Distributed dynamic map fusion via federated learning for intelligent networked vehicles,'' in \emph{2021 IEEE International conference on Robotics and Automation (ICRA)}.\hskip 1em plus 0.5em minus 0.4em\relax IEEE, 2021, pp. 953--959.

\bibitem{lossy}
J.~Li, R.~Xu, X.~Liu, J.~Ma, Z.~Chi, J.~Ma, and H.~Yu, ``Learning for vehicle-to-vehicle cooperative perception under lossy communication,'' \emph{IEEE Transactions on Intelligent Vehicles}, vol.~8, no.~4, pp. 2650--2660, 2023.

\bibitem{guang1}
C.~Liu, Y.~Chen, J.~Chen, R.~Payton, M.~Riley, and S.-H. Yang, ``Cooperative perception with learning-based v2v communications,'' \emph{IEEE Wireless Communications Letters}, vol.~12, no.~11, pp. 1831--1835, 2023.

\bibitem{guang2}
C.~Liu, J.~Chen, Y.~Chen, R.~Payton, M.~Riley, and S.-H. Yang, ``Self-supervised adaptive weighting for cooperative perception in v2v communications,'' \emph{IEEE Transactions on Intelligent Vehicles}, vol.~9, no.~2, pp. 3569--3580, 2024.

\bibitem{cooper}
Q.~Chen, S.~Tang, Q.~Yang, and S.~Fu, ``Cooper: Cooperative perception for connected autonomous vehicles based on 3d point clouds,'' in \emph{2019 IEEE 39th International Conference on Distributed Computing Systems (ICDCS)}.\hskip 1em plus 0.5em minus 0.4em\relax IEEE, 2019, pp. 514--524.

\bibitem{10398509}
C.~Lin, D.~Tian, X.~Duan, J.~Zhou, D.~Zhao, and D.~Cao, ``V2vformer: Vehicle-to-vehicle cooperative perception with spatial-channel transformer,'' \emph{IEEE Transactions on Intelligent Vehicles}, vol.~9, no.~2, pp. 3384--3395, 2024.

\bibitem{10265751}
H.~Yin, D.~Tian, C.~Lin, X.~Duan, J.~Zhou, D.~Zhao, and D.~Cao, ``V2vformer++: Multi-modal vehicle-to-vehicle cooperative perception via global-local transformer,'' \emph{IEEE Transactions on Intelligent Transportation Systems}, vol.~25, no.~2, pp. 2153--2166, 2024.

\bibitem{10891961}
------, ``Cuda-x: Unsupervised domain-adaptive vehicle-to-everything collaboration via knowledge transfer and alignment,'' \emph{IEEE Transactions on Neural Networks and Learning Systems}, vol.~36, no.~8, pp. 14\,144--14\,158, 2025.

\bibitem{syncnet}
Z.~Lei, S.~Ren, Y.~Hu, W.~Zhang, and S.~Chen, ``Latency-aware collaborative perception,'' in \emph{European Conference on Computer Vision}.\hskip 1em plus 0.5em minus 0.4em\relax Springer, 2022, pp. 316--332.

\bibitem{keypoints}
Y.~Yuan, H.~Cheng, and M.~Sester, ``Keypoints-based deep feature fusion for cooperative vehicle detection of autonomous driving,'' \emph{IEEE Robotics and Automation Letters}, vol.~7, no.~2, pp. 3054--3061, 2022.

\bibitem{amongus}
Y.~Li, Q.~Fang, J.~Bai, S.~Chen, F.~Juefei-Xu, and C.~Feng, ``Among us: Adversarially robust collaborative perception by consensus,'' in \emph{Proceedings of the IEEE/CVF International Conference on Computer Vision}, 2023, pp. 186--195.

\bibitem{doppler}
Y.~Zhao and S.-G. Haggman, ``Intercarrier interference self-cancellation scheme for ofdm mobile communication systems,'' \emph{IEEE transactions on Communications}, vol.~49, no.~7, pp. 1185--1191, 2001.

\bibitem{multipath}
W.~Dahech, M.~P{\"a}tzold, C.~A. Guti{\'e}rrez, and N.~Youssef, ``A non-stationary mobile-to-mobile channel model allowing for velocity and trajectory variations of the mobile stations,'' \emph{IEEE Transactions on Wireless Communications}, vol.~16, no.~3, pp. 1987--2000, 2017.

\bibitem{10681246}
\BIBentryALTinterwordspacing
Z.~Yue, J.~Wang, and C.~C. Loy, ``{ Efficient Diffusion Model for Image Restoration by Residual Shifting },'' \emph{IEEE Transactions on Pattern Analysis \& Machine Intelligence}, vol.~47, no.~01, pp. 116--130, Jan. 2025. [Online]. Available: \url{https://doi.ieeecomputersociety.org/10.1109/TPAMI.2024.3461721}
\BIBentrySTDinterwordspacing

\bibitem{diff_sr3}
C.~Saharia, J.~Ho, W.~Chan, T.~Salimans, D.~J. Fleet, and M.~Norouzi, ``Image super-resolution via iterative refinement,'' \emph{IEEE transactions on pattern analysis and machine intelligence}, vol.~45, no.~4, pp. 4713--4726, 2022.

\bibitem{diff_image}
B.~Xia, Y.~Zhang, S.~Wang, Y.~Wang, X.~Wu, Y.~Tian, W.~Yang, and L.~Van~Gool, ``Diffir: Efficient diffusion model for image restoration,'' in \emph{Proceedings of the IEEE/CVF International Conference on Computer Vision}, 2023, pp. 13\,095--13\,105.

\bibitem{cdiffse}
Y.-J. Lu, Z.-Q. Wang, S.~Watanabe, A.~Richard, C.~Yu, and Y.~Tsao, ``Conditional diffusion probabilistic model for speech enhancement,'' in \emph{ICASSP 2022-2022 IEEE International Conference on Acoustics, Speech and Signal Processing (ICASSP)}.\hskip 1em plus 0.5em minus 0.4em\relax IEEE, 2022, pp. 7402--7406.

\bibitem{diff_cddm}
T.~Wu, Z.~Chen, D.~He, L.~Qian, Y.~Xu, M.~Tao, and W.~Zhang, ``Cddm: Channel denoising diffusion models for wireless semantic communications,'' \emph{IEEE Transactions on Wireless Communications}, 2024.

\bibitem{10816175}
M.~Letafati, S.~Ali, and M.~Latva-Aho, ``Conditional denoising diffusion probabilistic models for data reconstruction enhancement in wireless communications,'' \emph{IEEE Transactions on Machine Learning in Communications and Networking}, vol.~3, pp. 133--146, 2025.

\bibitem{10674003}
------, ``Diffusion model-aided data reconstruction in cell-free massive mimo downlink: A computation-aware approach,'' \emph{IEEE Wireless Communications Letters}, vol.~13, no.~11, pp. 3162--3166, 2024.

\bibitem{diff_bit}
E.~Grassucci, S.~Barbarossa, and D.~Comminiello, ``Generative semantic communication: Diffusion models beyond bit recovery,'' \emph{arXiv preprint arXiv:2306.04321}, 2023.

\bibitem{diff_commin}
J.~Chen, D.~You, D.~G{\"u}nd{\"u}z, and P.~L. Dragotti, ``Commin: Semantic image communications as an inverse problem with inn-guided diffusion models,'' in \emph{ICASSP 2024-2024 IEEE International Conference on Acoustics, Speech and Signal Processing (ICASSP)}.\hskip 1em plus 0.5em minus 0.4em\relax IEEE, 2024, pp. 6675--6679.

\bibitem{ssd}
W.~Liu, D.~Anguelov, D.~Erhan, C.~Szegedy, S.~Reed, C.-Y. Fu, and A.~C. Berg, ``Ssd: Single shot multibox detector,'' in \emph{Computer Vision--ECCV 2016: 14th European Conference, Amsterdam, The Netherlands, October 11--14, 2016, Proceedings, Part I 14}.\hskip 1em plus 0.5em minus 0.4em\relax Springer, 2016, pp. 21--37.

\bibitem{PointNet}
R.~Q. {Charles}, H.~{Su}, M.~{Kaichun}, and L.~J. {Guibas}, ``Pointnet: Deep learning on point sets for {3D} classification and segmentation,'' in \emph{Proc. IEEE Conf. Comput. Vision Pattern Recognit. (CVPR)}, 2017, pp. 77--85.

\bibitem{8237586}
T.-Y. Lin, P.~Goyal, R.~Girshick, K.~He, and P.~Dollár, ``Focal loss for dense object detection,'' in \emph{2017 IEEE International Conference on Computer Vision (ICCV)}, 2017, pp. 2999--3007.

\bibitem{winner2}
\BIBentryALTinterwordspacing
P.~Kyosti, ``Winner ii channel models,'' \emph{IST, Tech. Rep. IST-4-027756 WINNER II D1.1.2 V1.2}, 2007. [Online]. Available: \url{https://cir.nii.ac.jp/crid/1571698600936523008}
\BIBentrySTDinterwordspacing

\bibitem{li2020practical}
W.~Li, Q.~Zhu, C.-X. Wang, F.~Bai, X.~Chen, and D.~Xu, ``A practical non-stationary channel model for vehicle-to-vehicle mimo communications,'' in \emph{2020 IEEE Wireless Communications and Networking Conference (WCNC)}.\hskip 1em plus 0.5em minus 0.4em\relax IEEE, 2020, pp. 1--6.

\bibitem{channel1}
Q.~Zhu, H.~Li, Y.~Fu, C.-X. Wang, Y.~Tan, X.~Chen, and Q.~Wu, ``A novel 3d non-stationary wireless mimo channel simulator and hardware emulator,'' \emph{IEEE Transactions on Communications}, vol.~66, no.~9, pp. 3865--3878, 2018.

\bibitem{channel2}
Q.~Zhu, Y.~Yang, X.~Chen, Y.~Tan, Y.~Fu, C.-X. Wang, and W.~Li, ``A novel 3d non-stationary vehicle-to-vehicle channel model and its spatial-temporal correlation properties,'' \emph{IEEE access}, vol.~6, pp. 43\,633--43\,643, 2018.

\bibitem{ddpm}
J.~Ho, A.~Jain, and P.~Abbeel, ``Denoising diffusion probabilistic models,'' \emph{Advances in neural information processing systems}, vol.~33, pp. 6840--6851, 2020.

\bibitem{v2v4real}
R.~Xu, X.~Xia, J.~Li, H.~Li, S.~Zhang, Z.~Tu, Z.~Meng, H.~Xiang, X.~Dong, R.~Song \emph{et~al.}, ``V2v4real: A real-world large-scale dataset for vehicle-to-vehicle cooperative perception,'' in \emph{Proceedings of the IEEE/CVF Conference on Computer Vision and Pattern Recognition}, 2023, pp. 13\,712--13\,722.

\end{thebibliography}

\end{document}